\documentclass{article}
\usepackage[utf8]{inputenc}
\usepackage[T1]{fontenc}
\usepackage{nicefrac}
\usepackage{algorithmic}
\usepackage[ruled,vlined]{algorithm2e}
\usepackage{amsmath,amsfonts,amssymb}
\usepackage{booktabs}
\usepackage{graphicx}
\usepackage{hyperref}
\usepackage{microtype}
\usepackage{natbib}
\usepackage{subcaption}
\newcommand{\kvector}{\mathbf{k}}

\usepackage{PRIMEarxiv}

\newcommand{\ins}[1]{#1}
\newcommand{\del}[1]{}
\newcommand{\sep}{\enspace\textbullet\enspace}
\newcommand{\linkref}[2][]{#1#2}
\newenvironment{imageonly}{}{}
\newcommand{\tbl}[2]{\caption{#1}#2}
\newcommand{\colrule}{\midrule}
\newcommand{\botrule}{\bottomrule}
\newcommand{\PrintCredit}{}
\newenvironment{coi}{}{}
\newenvironment{ack}{}{}
\newenvironment{appgroup}{}{}
\newcommand{\appsection}[1]{\section{#1}}

\title{Binned \del{S}\ins{s}pectral \del{P}\ins{p}ower \del{L}\ins{l}oss for \del{I}\ins{i}mproved \del{P}\ins{p}rediction of \del{C}\ins{c}haotic \del{S}\ins{s}ystems\protect\footnotemark}
\author{
	{\normalfont Dibyajyoti Chakraborty$^{1}$}
	\and
	Arvind T. Mohan$^{2}$
	\and
	Romit Maulik$^{3,4}$\\[0.5em]
	\normalfont $^{1}$ Pennsylvania State University, University Park, State College, PA, USA\\
	\normalfont $^{2}$ Los Alamos National Laboratory, Los Alamos, NM, USA\\
	\normalfont $^{3}$ Purdue University, West Lafayette, IN, USA\\
	\normalfont $^{4}$ Argonne National Laboratory, Lemont, IL, USA\\[0.25em]
	\normalfont \texttt{rmaulik@psu.edu}
}
\date{}

\begin{document}

\maketitle
\footnotetext{Published as: Chakraborty, D., Mohan, A. T., \& Maulik, R. (2026). \textit{Binned spectral power loss for improved prediction of chaotic systems}. \textit{Journal of Computational Physics}, 114866.\\
 \textbf{Sample codes:} \url{https://github.com/ISCLPurdue/bsp_chaos}.}

\begin{abstract}
	Forecasting multiscale chaotic dynamical systems, such as turbulent flows, with deep learning remains a formidable challenge due to the spectral bias of neural networks, which hinders the accurate representation of fine-scale structures in long-term predictions. This issue is exacerbated when models are deployed autoregressively, leading to compounding errors and instability. In this work, we introduce a novel approach to mitigate the spectral bias, which we call the Binned Spectral Power (BSP) Loss. The BSP loss is a frequency-domain loss function that adaptively weighs errors in predicting both larger and smaller scales of the dataset. Unlike traditional losses that focus on pointwise misfits, our BSP loss explicitly penalizes deviations in the energy distribution across different scales, promoting stable and physically consistent predictions. We demonstrate that the BSP loss mitigates the well-known problem of spectral bias in deep learning. We further validate our approach for the data-driven high-dimensional time-series forecasting of a range of benchmark chaotic systems, which are typically intractable due to spectral bias, culminating in experiments on canonical turbulent flow benchmarks. Our results demonstrate that the BSP loss significantly improves the stability and spectral accuracy of neural forecasting models without requiring architectural modifications. By directly targeting spectral consistency, our approach paves the way for more robust deep learning models for long-term forecasting of chaotic dynamical systems.
\end{abstract}

\section*{Highlights}
\begin{itemize}
	\item This work introduces a novel Binned Spectral Power (BSP) loss function that preserves the energy distribution across different spatial scales of a deep learning forecast model for dynamical systems. Unlike traditional losses, BSP adaptively weighs errors in frequency space, promoting more stable and physically consistent predictions without requiring architectural changes.
	\item We provide theoretical insights and empirical evidence on how the BSP loss improves spectral fidelity in deep learning models.
	\item The BSP loss is validated on high-dimensional, multiscale chaotic systems, including 2D and 3D turbulent flows and the Kuramoto-Sivashinsky equation. The results demonstrate that models trained with BSP loss significantly improve predictive stability and spectral accuracy over long-term forecasting horizons, capturing fine-scale structures that are often lost with standard loss functions.
	\item Our method not only improves the accuracy of the forecast, but also preserves key physical invariants and statistical properties of turbulent flows, such as the probability density functions of velocity, vorticity, and kinetic energy of the turbulence, outperforming baseline models.
\end{itemize}

\keywords{Surrogate \del{M}\ins{m}odeling\sep  Chaotic \del{S}\ins{s}ystems\sep  Deep \del{L}\ins{l}earning}
	
	\section{Introduction  }\label{Xsec1-1}
	The improved forecasting of complex nonlinear dynamical systems is of vital importance to several real-world applications, such as in engineering \citep{kong2022digital}, geoscience \citep{sun2024probabilistic}, public health \citep{wang2021bridging}, and beyond. Frequently, the accurate modeling of such systems is complicated by their multiscale nature and chaotic behavior. Physics-based models for such systems are generally described as partial differential equations (PDEs), the numerical solutions of which require significant computational effort. For instance, the presence of multiscale behavior requires very fine spatial and temporal resolutions, when numerically solving such PDEs, which can be severely limiting for real-time forecasting tasks \citep{harnish2021multiresolution}. Chaotic systems also require the assessment of statistics using ensembles of simulations, adding significant costs. This is one of the key bottlenecks in a variety of applications in earth sciences, energy engineering, and aeronautics.
	
	One approach to addressing the aforementioned challenges is through the use of data-driven methods for learning the temporal evolution of such systems. In such methods, function approximation techniques such as neural networks \citep{cybenko1989approximation,mcculloch1943logical}, Gaussian processes \citep{santner2003design}, and neural operators \citep{chen1995universal}, among others, are utilized to learn the map between subsequent time-steps from training data. Subsequently, these trained models are deployed autoregressively to perform roll-out forecasts for dynamics into the future. This approach holds particular promise for systems where large volumes of data are available from open-sourced simulations or observations. Recently, this approach to forecasting has been applied with remarkable success to dynamical systems emerging in applications such as weather \citep{bi2022pangu,lam2022graphcast,pathak2022fourcastnet,nguyen2023scaling}, climate \citep{guan2024lucie,watt2023ace,ruhling2024dyffusion}, nuclear fusion \citep{mehta2021neural,burby2020fast,li2024surrogate}, renewable energy \citep{sun2019short,wang2019review}, etc.
	
	However, purely data-driven forecast models suffer from a common limitation that degrades their performance in comparison with physics-based solvers. This pertains to an inability to capture the information at smaller scales in the spatial domain of the dynamical system \citep{bonavita2024some,olivetti2024data,pasche2025validating,mahesh2024huge}. In the spectral space, these refer to the energy associated with higher wavenumbers. Consequently, data-driven models may be over or under-dissipative during autoregressive predictions, which eventually cause a significant disagreement with ground-truth and in worse-case scenarios, leading to completely non-physical behavior \citep{chattopadhyay2023long}. These errors are commonly understood to be caused by so-called \textit{spectral biases} \citep{rahaman2019spectral}, defined by the tendency of a neural network trained on a typical mean-squared-error loss function to optimize the larger wavenumbers first while training. This phenomenon has been observed across a variety of neural network architectures, across differing applications, like generative adversarial networks \citep{schwarz2021frequency,chen2021ssd}, transformers \citep{bhattamishra2022simplicity}, state space models \citep{yu2024tuning}, physics-informed neural networks \citep{chai2024overcoming}, Kolmogorov-Arnold networks \citep{wang2024expressiveness}, etc.
	
	\textbf{Related Works :} Significant research has focused on addressing the challenges of difficulty in capturing high-frequency structures \citep{karniadakis2021physics,lai2024machine,chakraborty2024divide,chen2024physics}. A major direction of work involves architectural innovations in neural networks aimed at mitigating spectral bias and improving the resolution of fine-scale features. For instance, \cite{tancik2020fourier} introduces Fourier feature mappings to enhance fully connected networks, while the Hierarchical Attention Neural Operator (HANO) proposed by \cite{liu2024mitigating} leverages multilevel representations with self-attention and local aggregation to capture multiscale dependencies. Similarly, diffusion models have shown promise by modeling the forecast as a sample from a learnable stochastic process \citep{gao2023implicit,oommen2024integrating,luo2023image,whittaker2024turbulence}. PDE-Refiner \citep{lippe2023modeling} progressively refines predictions to capture both dominant and weak frequency modes. Gestalt autoencoders \citep{liu2023devil} enhance reconstruction in both spatial and spectral domains, while frequency-aware training strategies such as dynamic spectral weighting have been proposed to prioritize specific wavenumber bands \citep{lin2023catch}. Multiscale neural approximations and hierarchical discretization frameworks have also been used to improve fine-scale information exchange and prediction quality \citep{barwey2023multiscale,wang2020multi,liu2020multi,khodakarami2025mitigating}. Some new approaches propose choices for hyperparameters or data processing to improve the quality of the predictions \citep{cai2024batch}. Using statistical measures in the optimization objective \citep{schiff2024dyslim,wu2020enforcing} and reducing error accumulation by integrating recurrent neural networks (RNNs) with neural operators \citep{michalowska2024neural} have also been previously explored. Another direction is to use hybrid techniques which combine numerical solvers with neural networks to improve physical consistency\citep{shankar2023differentiable,zhang2024blending,chen2025neural,geneva2020modeling,khodkar2021data}. Despite their effectiveness, many of these methods involve complex architectural designs or heavy computational overhead.
	
	We aim to address the following open question: \textit{What is a universally implementable augmentation for the deep learning of chaotic dynamical systems so that spectral bias may be mitigated and invariant statistics be preserved with minimal impact on computational cost?} In this work, we propose such an approach, with a particular focus on its application in forecasting turbulent flows.
	
	\textbf{Contributions :} The contributions of this paper are as follows: First, we introduce the Binned Spectral Power (BSP) Loss, a novel approach to address the spectral bias of arbitrary neural forecasting models. By focusing on preserving the distribution of energy across different spatial scales instead of relying solely on pointwise comparisons, our method enhances the stability and quality of long-term predictions. Second, our proposed framework is architecture agnostic, easily deployable, and requires minimum additional hyperparameter tuning. This ensures that our approach remains broadly applicable, computationally feasible, and adaptable to a variety of dynamical systems. Third, we show that the BSP loss can actually mitigate the spectral bias using a synthetic example from \cite{rahaman2019spectral}. Fourth, we further examine the effectiveness of our method through extensive testing on the forecasting of the following complex and high-dimensional chaotic systems: Kolmogorov flow \citep{obukhov1983kolmogorov}, a 2D benchmark for chaotic systems used for various studies \citep{kochkov2021machine}, a high Reynolds number flow over NACA0012 airfoil \citep{towne2023database} and the 3D homogeneous isotropic turbulence \citep{mohan2020spatio}. Our results indicate that the proposed loss function significantly improves both predictive stability and spectral accuracy, mitigating common limitations of deep learning models in capturing fine-scale structures over long forecasting horizons.
	
	\section{Background}\label{Xsec2-2}
	We consider an operator \( G \) that maps one timestep of the state \( x \) of a dynamical system to the next. This operator can be viewed as the \textit{optimal} data-driven process that bypasses the direct solution of the governing differential equation for each timestep, effectively describing the system's dynamics. {The evolution of the\del{ the} state at time \( t \) is given as $x_t = G(x_{t-1}) = G(G(G(\ldots G(x_0)))) = G^t(x_0)$.} The operator \( G \) can be approximated using a neural network model \( F_\phi(x) \), parameterized by learnable variables \( \phi \). Such an approximation is backed by the universal approximation theorem for operators \citep{chen1995universal}. These parameters of \( F_\phi(x) \) are optimized by minimizing the discrepancy from the ground truth data (indexed discretely by \( j \)) using a one-step loss function defined as:
	\begin{equation}\label{MSE-1}
		L = \mathbb{E}_j\left[\left\|F_{\phi}(x_j) - G(x_j)\right\|^2\right].
	\end{equation}
	A commonly employed multi-rollout loss function \citep{keisler2022forecasting}, \( L_R \), utilized in training many state-of-the-art models, is defined as:
	\begin{equation}\label{MSE-2}
		L_m = \mathbb{E}_j \left[\sum_{t=1}^{t=m} \left\| \gamma(t) \big(F_{\phi}^t(x_j) - G^t(x_j)\big)\right\|^2\right],
	\end{equation}
	where \( m \) denotes the number of rollouts included during training, and \( \gamma(t) \)
	is a hyperparameter that assigns diminishing weights to errors in trajectories further along in time \citep{kochkov2023neural}. It has an effect similar\footnote{\footnotesize Although the discount factor in RL is unrelated directly to the \( \gamma(t) \) used here, there might be interesting theoretical connections which we leave for future exploration.} to the discount factor used in reinforcement learning(RL) \citep{amit2020discount}. Furthermore, to enhance computational efficiency and improve stability, the \textit{Pushforward Trick}, introduced in \cite{brandstetter2022message}, is often used. This approach reduces computational overhead by detaching the computational graph at intermediate rollouts. However, such methods alone cannot address either the phenomenon of spectral bias of neural networks nor stability in long-term rollouts \citep{chakraborty2024divide,schiff2024dyslim}.
	
	\subsection{Spectral \del{B}\ins{b}ias in \del{D}\ins{d}eep \del{L}\ins{l}earning}\label{Xsec3-2.1}
	\citet{rahaman2019spectral} showed that a combination of the theoretical properties of gradient descent optimization, the architecture of neural networks, and the nature of function approximation in high-dimensional spaces causes the network to learn lower frequencies faster and more effectively. In our case, for $N$ samples in a training batch, \linkref[\del{Equation}\ins{Eq.}]{\ref{MSE-1}} can be approximated by $L_1$ as follows, where subscript $1$ signifies one step MSE loss.
	\begin{equation}\label{MSE-1-act}
		L_1 = \frac{1}{N}\sum_{j=0}^N\left\|F_{\phi}(x_j) - G(x_j)\right\|^2.
	\end{equation}
	{The gradient of this loss function with respect to parameters $\phi$ is given by $\nabla_\phi L_1 = \frac{2}{N}\sum_{j=0}^N\left(F_{\phi}(x_j) - G(x_j)\right)\nabla_\phi F_\phi(x_j)$, which may be used in a gradient descent update step as $\phi_{k+1} = \phi_k - \alpha \nabla_\phi L_1$, where $\alpha$ is the learning rate.} Intuitively, gradient descent naturally favors changes that yield the most substantial reduction in loss early in training. In the spectral space, this is reflected in the components that have higher values in the Fourier series representation of $F_\phi$ \citep{oommen2024integrating}. This causes the lower frequencies to be learned first, which correspond to global patterns that tend to dominate the error landscape in the initial phases of training. For more details, readers are directed to \linkref[Section]{\protect\ref{Xsec6-3}} in \cite{rahaman2019spectral} and \linkref[Section]{\protect\ref{Xsec9-4.1}} in \cite{oommen2024integrating}.
	
	\subsection{Energy \del{S}\ins{s}pectrum}\label{Xsec4-2.2}\label{Sec:Energy_Spec}
	{The energy spectrum $E(k)$ characterizes the distribution of energy among different frequency or wavenumber components \citep{kolmogorov1941local}. In our work, the Fourier Transform is always taken spatially. However, we use the terms frequency and wavenumber interchangeably henceforth. For an arbitrary field $u(x)$ (can be $F_\phi(x)$ or $G(x)$ from \linkref[\del{Equation}\ins{Eq.}]{\ref{MSE-1}}) in a periodic domain of length $L$, the \textit{Fourier transform} $\mathcal{F}$ is defined as $\hat{u}(k) = \mathcal{F}(u(x)) = \frac{1}{L} \int_0^L u(x) e^{-ikx} dx$, where $\hat{u}(k)$ represents the spectral coefficients corresponding to wavenumber $k$.}
	
	{For higher-dimensional fields $u(x,y,t)$ or $u(x,y,z,t)$, the Fourier transform is extended to multiple dimensions, and the energy density is computed by summing over all wavevectors of the same magnitude: $E(k) = \frac{1}{2} \sum_{|\mathbf{k}| = k} |\hat{u}(\mathbf{k})|^2$, where $\mathbf{k} = (k_x, k_y, k_z)$ is the wavevector, and summation is performed over spherical shells in Fourier space. In computational settings, we often work with discretized fields defined on a uniform grid. The discrete Fourier transform (DFT) is used to approximate the energy spectrum: $\hat{u}(\mathbf{k}) = \frac{1}{N} \sum_{n=0}^{N-1} u_n e^{-i 2\pi kn/N}$, where $N$ is the number of grid points.
		For handling discrete wavenumbers in computational grids, binning helps to efficiently average the energy over wavenumber shells, ensuring a smooth representation of the spectrum. The magnitude of each wavenumber $k$ is given as
		\begin{equation}\label{wavenumber_mag}
			k = \sqrt{k_x^2+k_y^2+k_z^2}
		\end{equation}
		The bins can be logarithmically or linearly spaced. In our experiments, we use linearly spaced bins for computing the energy contributions into wavenumber shells as:
		\begin{equation}\label{eq:binned_specra}
			E(k) = \sum_{k - \Delta k/2 \leq |\mathbf{k}| < k + \Delta k/2} \frac{1}{2} |\hat{u}(\mathbf{k})|^2,
		\end{equation}
		where $\Delta k$ is the width of the bin. In several scenarios, a major portion of the energy is stored in the lower wavenumbers, highlighted by the rapid decay of their energy spectrum. However, in complex real-world systems, the energy spectrum typically exhibits a slow decay, preserving substantial energy and valuable information at higher wave numbers. For example, in weather data, the small and intermediate scale details correspond to anomalies like initial phases of storms \citep{ritchie1997scale}, especially in a model with coarser grids.}
	
	\subsection{Regularization in \del{F}\ins{f}ourier \del{S}\ins{s}pace}\label{Xsec5-2.3}
	An intuitive solution to the problem of capturing the fine scales can be to penalize the mismatch of the Fourier transform of the model outputs from the ground truth \citep{chattopadhyay2024oceannet,guan2024lucie,kochkov2023neural}. {This is typically done by a regularization in the Fourier space, such as
		\begin{equation}\label{MSE-Fspace}
			L_f = \frac{1}{N} \sum_{j=0}^{N-1} \, \sum_{k} w_k \, \left| \mathcal{F} \left( F_{\phi}(x_j) - G(x_j) \right) \right|_k^2.
		\end{equation}
		where $\mathcal{F}$ is the Fourier transform, and $w_k$ is a hyperparameter used to weigh or truncate specific modes. It is evident that \linkref[\del{Equation}\ins{Eq.}]{\ref{MSE-Fspace}} will also be heavily biased towards the larger values in the Fourier spectrum, which typically correspond to the lower frequency modes. For example, if $w_k=1$, the effect of \linkref[\del{Equation}\ins{Eq.}]{\ref{MSE-Fspace}} is the same as the loss function in \linkref[\del{Equation}\ins{Eq.}]{\ref{MSE-1-act}}. To overcome this, \cite{chattopadhyay2024oceannet} used a cutoff to empirically ignore some of the lower frequencies\del{.}\cite{guan2024lucie}\ins{.} used a mean absolute error in the tendency space after Fourier transform to obtain better performance. However, for higher frequencies with extremely low contributions, it is not judicious to try to match them exactly in a point-wise manner. This is demonstrated by our experiments in later sections. Another version of this loss function where $w_k = {(1+{|k|}^2)}^s$ is called the Sobolev Loss \citep{li2021markov,czarnecki2017sobolev}. It shows promise in PDE applications as the Sobolev norm corresponds to certain physical quantities (e.g., energy, enstrophy). We compare against this loss function in further sections. However, we note that the weight in the Sobolev loss is fixed to $k^2$ and is not determined by the distribution of energy in different scales of the training data. In the following section, we come up with a new strategy to solve the mentioned problems without modifying the network architecture or incurring a heavy cost during training and inference.}
	
	\section{Methodology}\label{Xsec6-3}
	
	\begin{algorithm}[!h]
		\small
		\caption{Binned Spectral Power (BSP) Loss Computation.}
		\label{alg:spectral_loss}
		
		\begin{imageonly}
			\begin{algorithmic}
				\REQUIRE Predicted data $u_j$, Target data $v_j \in \mathbb{R}^{C \times H \times W \times \del{...}\ins{\ldots }}$, a small positive constant $\epsilon$
				\REQUIRE Number of wavenumber bins $N_k$, and method to define bin $i$ (e.g., linear : 0\del{-}\ins{--}1, 1\del{-}\ins{--}2,.\del{.})
				\REQUIRE Non-negative weights $\lambda_i$ for each bin $i=1, \dots, N_k$
				\ENSURE Spectral Loss $L_\text{spec}^{(j)}$
				\\
				\textit{\# N-D Spatial Fourier Transform}
				\STATE $\hat{u} \leftarrow \mathcal{F}(u_j)$, $\hat{v} \leftarrow \mathcal{F}(v_j)$
				\\
				\textit{\# Energy per mode $(c,\kvector)$}
				\STATE $E_u \leftarrow \frac{1}{2} |\hat{u}|^2$, $E_v \leftarrow \frac{1}{2} |\hat{v}|^2$
				\\
				\textit{\# Wavenumber magnitude}
				\STATE $k \leftarrow \sqrt{\kvector_x^2 + \kvector_y^2 + \dots}$ \hfill
				\\
				\textit{\# Avg. $E_u(c,\kvector)$ in bin and scale by $\lambda_i$}
				\FOR{$i = 1$ to $N_k$}
				\STATE {$E_u^\text{bin}(c, i) \leftarrow \left(\frac{1}{N_i} \sum_{k \in \text{bin}_i} \lambda_i \cdot E_u(c, \kvector)   \right)$\hfill }
				\STATE { $E_v^\text{bin}(c, i) \leftarrow \left(\frac{1}{N_i} \sum_{k \in \text{bin}_i} \lambda_i \cdot E_v(c, \kvector) \right)$ \hfill }
				\ENDFOR
				\\
				\textit{\# Final loss computation}
				\STATE $L_\text{{BSP}}^{(j)} \leftarrow {\frac{1}{N_k C}} \sum_{c=1}^C \sum_{i=1}^{N_k} \left( 1 - \frac{E^{\text{bin}}_u(c,i)+ \epsilon}{E^{\text{bin}}_v(c,i) + \epsilon} \right)^2$ \hfill
			\end{algorithmic}
		\end{imageonly}
		
	\end{algorithm}
	
	We introduce a novel Binned Spectral Power (BSP) loss function mentioned in \linkref[Algorithm]{\ref{alg:spectral_loss}}. This is designed to evaluate discrepancies between predicted and target data fields by comparing their spatial energy spectra at different scales. We reuse the concept of energy spectrum mentioned in \linkref[Section]{\ref{Sec:Energy_Spec}}. First, the predicted and target samples are transformed into the wavenumber domain using the Fourier transform. The magnitudes of energy components are computed by squaring the Fourier coefficients. The wavenumber magnitudes are then computed using \linkref[\del{Equation}\ins{Eq.}]{\ref{wavenumber_mag}} to group spatial frequency components into scalar values. The energy components are binned by wavenumber ranges, averaging the energy within each bin $E^{bin}$ using \linkref[\del{Equation}\ins{Eq.}]{\ref{eq:binned_specra}}. Here every bin $(k)$ is defined as $(k - \Delta k/2 )\leq |\mathbf{k}| < (k + \Delta k/2)$. The BSP loss is calculated by comparing the binned energy spectra of the predicted and target samples.
	
	Unlike traditional loss functions like Mean Squared Error (MSE), which operate point-wise in the physical domain, the BSP loss provides a robust learning of the various scales in the data, as explained in the following. To ensure the accurate capturing of different scales we aim to get the ratio of the energy in different bins close to identity. This squared relative error loss is successful in providing equal weights to the energy component at all wavenumber bins. The BSP Loss is defined as:
	\begin{equation}
		L_\text{BSP}(u,v) = {\frac{1}{N_k C}} \sum_{c=1}^C \sum_{i=1}^{N_k} \left( 1 - \frac{E^{\text{bin}}_u(c,i)+ \epsilon}{E^{\text{bin}}_v(c,i) + \epsilon} \right)^2
		\label{Xeqn7-7}
	\end{equation}
	where $N_k$ is the number of bins, $i$ refers to a specific bin spanning a range of wavenumbers, and $C$ is the number of features  (channels) in input $u$ and target $v$. $\epsilon$ is used to eliminate the effect of extremely small values in $E^{bin}$. {The hyper-parameter $ \lambda_i$ (see \linkref[Algorithm]{\ref{alg:spectral_loss}}) is used to variably weight different bins based on the requirements of the application. In most cases, this can be set to unity -- however, special treatment may be needed for specific examples (see Experiment section).} In practice, we recommend setting $u$ to the predicted values and $v$ to the true values in $L_{\text{BSP}}$, such that the neural network outputs appear in the numerator. This formulation yields more stable gradients and is computationally simpler to differentiate. The algorithm can be written in a differentiable programming language to efficiently compute the gradients required to minimize the BSP loss. A differentiable histogram can also be used to efficiently perform the binning using performant libraries like Jax \citep{jax2018github}.
	
	The BSP loss can be combined with the multi-step rollout loss given in \linkref[\del{Equation}\ins{Eq.}]{\ref{MSE-2}} for short term accuracy, long term stability, and spectral bias mitigation.
	\begin{equation}\label{Loss eqn}
		L_R^* = \mathbb{E}_j \left[\sum_{t=1}^{t=m} \left\| \gamma(t) \big(F_{\phi}^t(x_j) - G^t(x_j)\big)\right\|^2 +  \mu L_\text{BSP}^{(j,t)}\right]
	\end{equation}
	where
	\begin{equation}
		L_\text{BSP}^{(j,t)} = L_\text{BSP}(F_{\phi}^t(x_j),G^t(x_j))
		\label{Xeqn9-9}
	\end{equation}
	is the BSP loss at $t^{th}$ autoregressive rollout step of the model, and $\mu$ is a hyper-parameter that is used to weigh the two loss terms differently. The gradient of the BSP loss is
	\begin{equation}
		\begin{aligned}
			\nabla_\phi L_\text{BSP} =
			\frac{-2}{N} \sum_{j=1}^N \frac{1}{N_k} \sum_{i=1}^{N_k} \sum_{c=1}^C  \left( 1 -
			\frac{E^{bin}_F(c,i) + \epsilon}{E^{bin}_G(c,i) + \epsilon} \right)
			\frac{\nabla_\phi E^{bin}_F(c,i)}{E^{bin}_G(c,i) + \epsilon}
		\end{aligned}
		\label{Xeqn10-10}
	\end{equation}
	It can be shown, following a similar treatment as for the MSE Loss (also refer to \linkref[Section]{\protect\ref{Xsec9-4.1}} from \citet{oommen2024integrating}), that the ratio term present in the gradient of the BSP loss leads to equal importance to all ranges of the energy spectrum. However, combining the BSP loss with the mean square error loss gives slightly higher importance to the lower wavenumbers, which is desirable as they contain the maximum energy. The weight $\mu$ can be adjusted to compensate for this when needed. {A detailed mathematical reasoning on the training dynamics using BSP loss and its comparison with MSE loss is shown in \ref{BSP theory}.} \textit{The BSP loss mentioned henceforth is the combined MSE + BSP loss mentioned in \linkref[\del{Equation}\ins{Eq.}]{\ref{Loss eqn}}.}
	
	\subsection{Complexity}\label{Xsec7-3.1}
	The BSP loss introduces minimal computational overhead compared to baseline objectives. The additional cost scales linearly with batch size ($n_b$) and quasi-linearly with the state dimension ($d$). It is easily estimated by considering the cost of the FFT step and assuming a small number of frequency bins ($N_k \ll d$). As detailed in \linkref[Table]{\ref{Comp_table}}, BSP has lower time and space complexity than MMD \citep{schiff2024dyslim}, and is comparable to standard MSE and push-forward losses. Here, $d$ is the state dimension, $|\phi|$ the number of model parameters, $NN$ the cost of a network forward pass, $n_b$ the batch size, and $n_t$ the number of rollout steps. MSE$_1$ and MSE$_t$ refer to one-step and multi-step MSE losses, respectively, and Pfwd denotes the push-forward trick from \cite{brandstetter2022message}.
	
	\begin{table}
		\tbl{Time and space complexity of different objectives.}{%
			\begin{tabular}{lll}
				\toprule
				Objective & Cost $\mathcal{O}(\cdot)$ & Memory $\mathcal{O}(\cdot)$ \\
				\colrule
				MSE$_1$  & $n_b d + n_b NN$        & $n_b d + n_b |\phi|$ \\
				MSE$_t$  & $n_t n_b d + n_t n_b NN$ & $n_t n_b d + n_t n_b |\phi|$ \\
				Pfwd     & $n_b d + n_b NN$        & $n_b d + n_b |\phi|$ \\
				MMD      & $n_b^2\del{ }\ins{~}d + n_b NN$       & $n_b^2\del{ }\ins{~}d + n_b |\phi|$ \\
				BSP      & $n_b d \log d + n_b NN$  & $n_b d + n_b |\phi|$ \\
				\botrule
		\end{tabular}}
		\label{Comp_table}
	\end{table}
	
	\section{Experiments}\label{Xsec8-4}
	We test our proposed methodology for several benchmark problems. These experiments aim to test the capabilities of our proposed loss function in the context of preserving small scale structures when applied to high-dimensional dynamical systems using existing deep learning architectures.
	
	\subsection{Mitigating the \del{S}\ins{s}pectral \del{B}\ins{b}ias}\label{Xsec9-4.1}\label{SB_section}
	
	\begin{figure}
		\begin{center}
			\begin{minipage}[t]{0.43\linewidth}
				\centering
				\includegraphics[width=\linewidth]{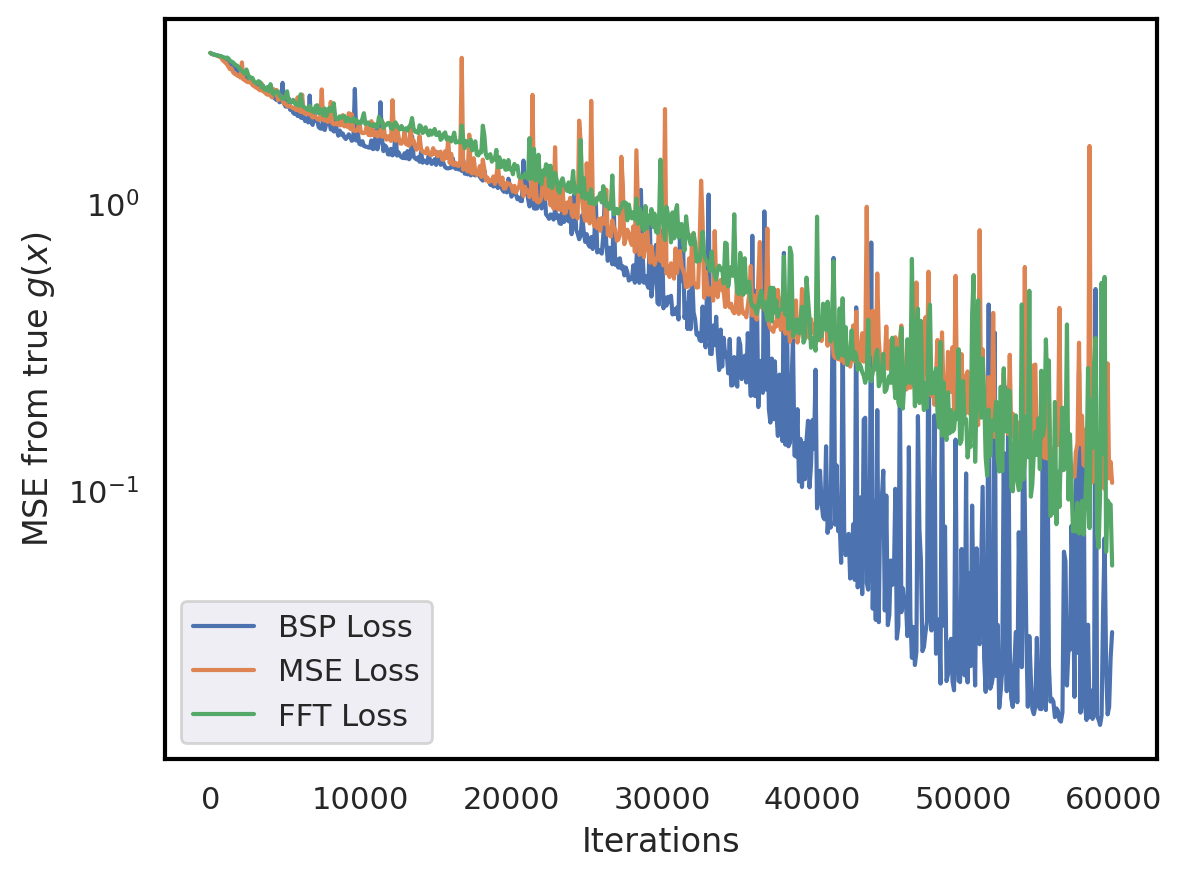}
			\end{minipage}
			\hfill
			\begin{minipage}[t]{0.56\linewidth}
				\centering
				\includegraphics[width=\linewidth]{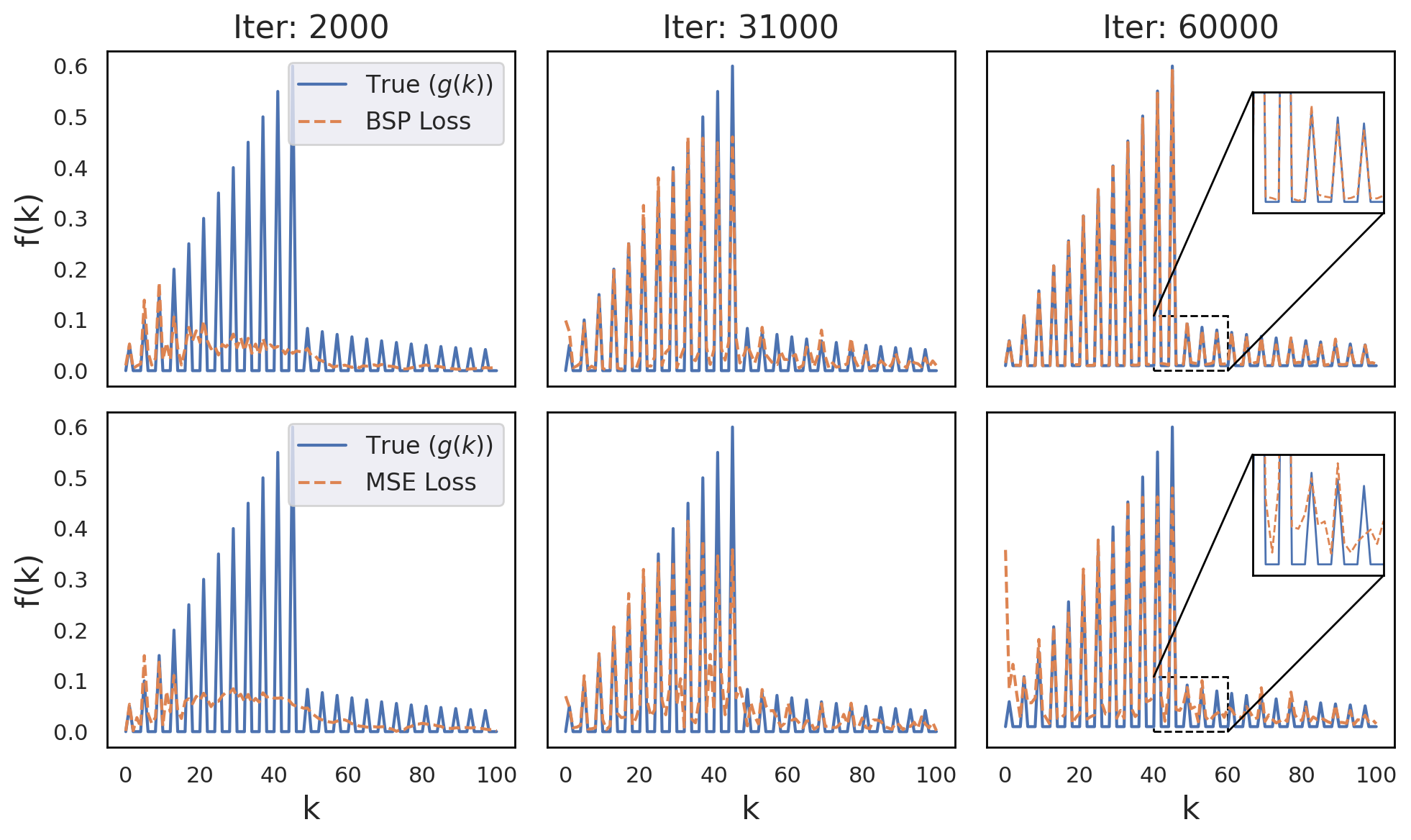}
			\end{minipage}
		\end{center}
		\caption{(left) MSE over training iterations for BSP Loss (blue), MSE (orange), and FFT Loss (green), showing faster convergence of BSP. (right) Frequency domain plot of predictions across training: BSP (top) recovers high-frequency components of $g(k)$ better than MSE (bottom).}
		\label{fig:spec_bias_combined_fft}
	\end{figure}
	
	We follow \cite{rahaman2019spectral} to evaluate the mitigation of spectral bias using BSP loss. A target function $g(x)= \sum_i A_i \sin(2\pi k_i x + \phi_i)$ is constructed as a sum of sinusoidal components with varying frequencies, amplitudes, and phases. A 6-layer ReLU network with 256 units per layer is trained to approximate $g(x)$ using 200 uniformly spaced samples over $[0,1]$. We compare models trained with standard MSE loss versus BSP loss. The target function $g : [0,1] \to \mathbb{R}$ is a weighted sum of sinusoids:
	\begin{equation}
		g(x) = \sum_{i} A_i \sin(2\pi k_i x + \phi_i),
		\label{Xeqn11-11}
	\end{equation}
	where $\kappa = (5, 10, \dots, 50)$ are the frequencies, amplitudes $\alpha = (A_1, \dots, A_n)$ vary smoothly from 0.08 to 1.2, and $\phi_i$ are uniformly sampled phases. The amplitudes rise to a peak and fall off, to highlight spectral bias in the learned function (see \linkref[\del{Figure}\ins{Fig.}]{\ref{fig:spec_bias_combined_fft}} (right)). We train a 6-layer ReLU network with 256 units per layer on 200 uniform samples over $[0,1]$, for \del{60,000}\ins{60,000} iterations. Two variants are compared: one trained with MSE loss and another with BSP loss. Since this is a 1D problem, the Fourier transform directly resolves the wavenumber content, so no binning is required. This leads to the simplified form of BSP loss:
	\begin{equation}
		L =  \left\|f_{\phi}(x) - g(x)\right\|^2 +  \mu \left[ 1 - \frac{\|\mathcal{F}(f_{\phi}(x))\| + \epsilon}{\|\mathcal{F}(g(x))\| + \epsilon} \right]^2,
		\label{Xeqn12-12}
	\end{equation}
	where $\mu = 5$ controls the strength of spectral alignment and $\epsilon = 1$ ensures numerical stability.
	
\begin{figure}[ht]
		
		\centerline{\includegraphics[width=0.8\linewidth]{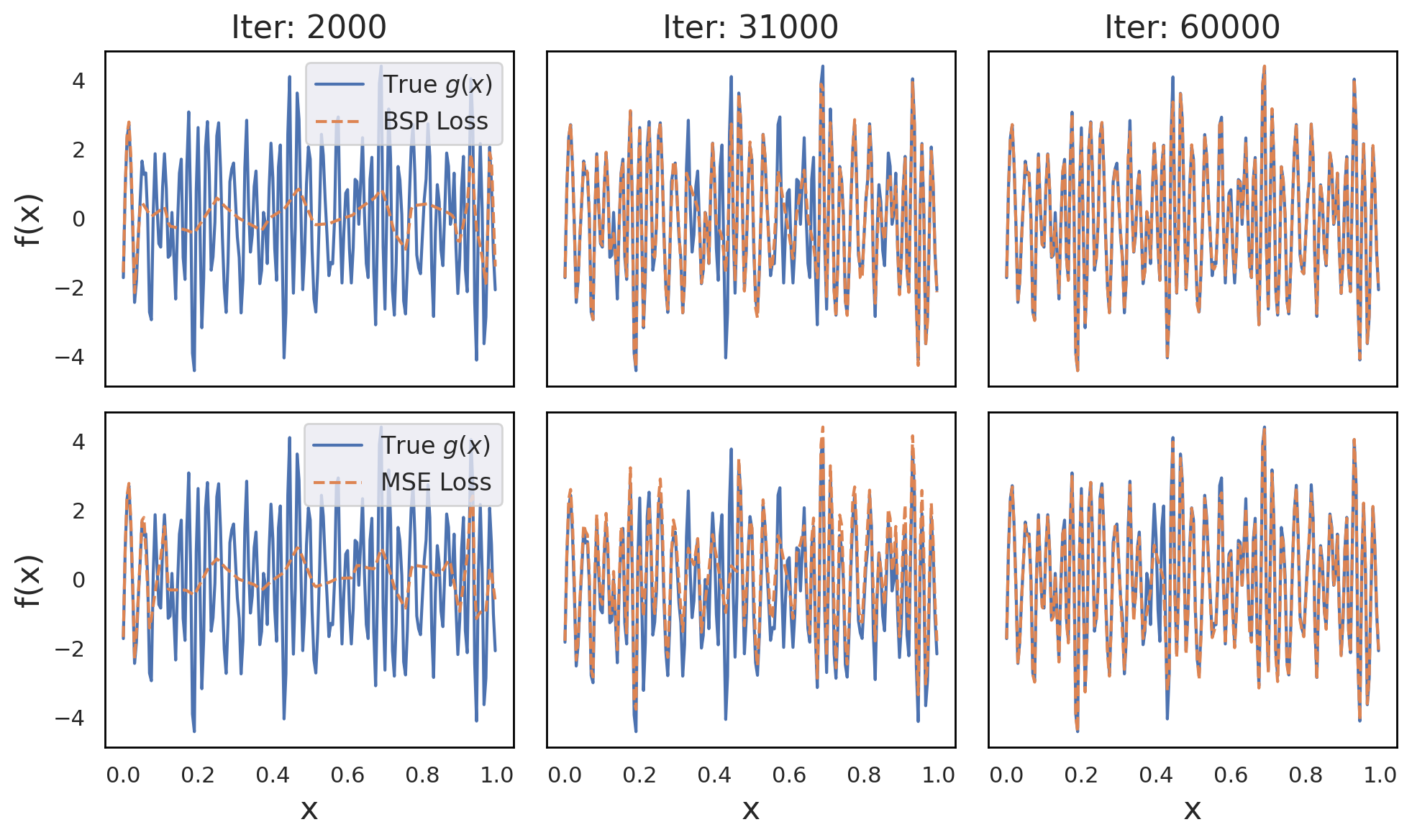}}
		\caption{Function approximation across training iterations. Top: BSP Loss; Bottom: MSE Loss. BSP better captures sharp transitions and high-frequency modes early in training.}
		\label{fig:spec_bias_func}
	\end{figure}

	In higher-dimensional problems, Fourier modes are typically grouped by isotropic wavenumber magnitude (see \linkref[\del{Equation}\ins{Eq.}]{\ref{wavenumber_mag}}), requiring binning across Cartesian shells. This is not needed here due to the 1D structure. \linkref[\del{Figure}\ins{Fig.}]{\ref{fig:spec_bias_func}} shows that a model trained with BSP Loss reconstructs the target function $g(x)$ more accurately than models trained with MSE Loss, especially during the initial phases of training.
	
	The impact of BSP Loss on function approximation and frequency learning is evident across the training iterations. The model trained with BSP Loss reconstructs the true function $g(x)$ with higher accuracy compared to those trained with MSE Loss, particularly in the earlier training stages (refer \linkref[\del{Figure}\ins{Fig.}]{\ref{fig:spec_bias_func}}). The advantage of BSP Loss is highlighted in \linkref[\del{Figure}\ins{Fig.}]{\ref{fig:spec_bias_combined_fft}} (right), where its Fourier Transform representations capture high-frequency components of the true function $g(k)$ more effectively than MSE Loss, which struggles to learn these components. Additionally, in \linkref[\del{Figure}\ins{Fig.}]{\ref{fig:spec_bias_combined_fft}} (left), we indicate the Mean Squared Error (MSE) throughout training iterations for the MSE loss, the BSP loss, and the FFT regularizer mentioned in \citep{chattopadhyay2024oceannet}. Although the FFT loss performs slightly better than just using the MSE loss, BSP is observed to outperform all of them, illustrating its ability to improve convergence. Additionall,y we would like to mention that we can not use the MMD loss here as it is a simple function approximation task and there is no concept of underlying invariant measure or attractor. These results collectively demonstrate that BSP Loss mitigates spectral bias and enhances function approximation by preserving the higher-frequency information in the learning process. Additionally, we also perform an ablation study for the hyperparameters in the BSP loss function, namely $\mu$ and $\epsilon$. From \linkref[Table]{\ref{ablation_table}}, it is observed that for all values of $\mu$ that we considered, the BSP loss consistently shows better performance by an order of magnitude from other baselines.
	
	\begin{table}
		\tbl{Comparison of mean square error at the end of optimization metrics for different values of $\mu$ for the Synthetic Experiment in \linkref[Section]{\ref{SB_section}}. The table compares models trained with MSE loss, BSP loss, and FFT loss \citep{chattopadhyay2024oceannet}. The MSE loss column is added for comparison, as it does not have the hyperparameter $\mu$. The best performing model is highlighted in bold.}{%
			\begin{tabular}{llll}
				\toprule
				${\mu}$ & {MSE} & {BSP} & {FFT} \\
				\colrule
				0.1   &      &0.206  $\pm$ 0.190        & 0.302 $\pm$ 0.213      \\
				1     &     &0.026 $\pm$ 0.011 & 0.081  $\pm$ 0.027     \\
				5     &  0.202$\pm$0.057 & \textbf{0.018 $\pm$ 0.007}& 0.226 $\pm$ 0.045      \\
				7.5   &      & 0.048  $\pm$ 0.033       & 0.260  $\pm$ 0.024     \\
				10    &      & 0.081  $\pm$  0.045      & 0.381  $\pm$  0.012    \\
				\botrule
		\end{tabular}}
		\label{ablation_table}
	\end{table}
	
	\subsection{Kuramoto-Shivashinsky \del{E}\ins{e}quation}\label{Xsec10-4.2}
	
	\begin{figure}
		
		\centerline{\includegraphics[width=\linewidth]{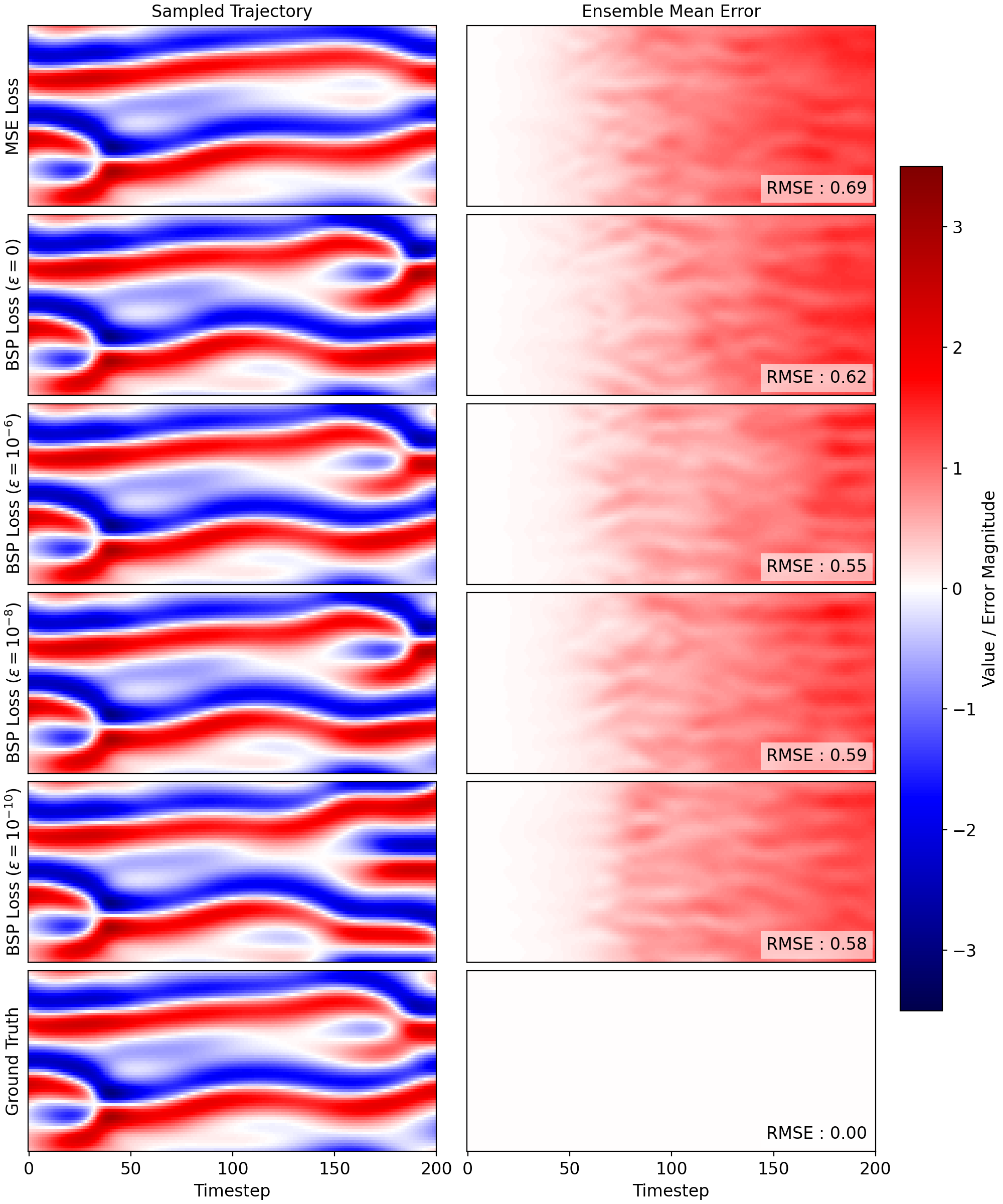}}
		\caption{Comparison of predicted trajectories (left) and ensemble mean absolute error (right) for models trained with different loss functions. Rows correspond to models trained with MSE loss and BSP loss with varying $\varepsilon \in \{0, 10^{-6}, 10^{-8}, 10^{-10}\}$, along with the ground truth (bottom row). BSP-trained models exhibit reduced forecast error, particularly for larger values of $\varepsilon$.}
		\label{fig:KS_image}
	\end{figure}
	
	The Kuramoto--Sivashinsky (KS) equation is a nonlinear partial differential equation that shows chaotic dynamics and is used as a benchmark for comparing forecast models \citep{lippe2023pde,li2021markov,jiang2023training,qu2022learning}. In one spatial dimension, it is given by:
	\begin{equation}
		\partial_t u + u \partial_x u + \partial_{xx} u + \partial_{xxxx} u = 0,
		\label{eq:ks}
	\end{equation}
	where \( u(x,t) \) represents the evolving field, typically taken to be periodic in space. The term \( u \partial_x u \) introduces nonlinearity, \( \partial_{xx} u \) accounts for linear instability, and the hyperviscous term \( \partial_{xxxx} u \) provides stabilizing dissipation. Despite its simple form, the KS equation exhibits spatiotemporal chaos and is often used as a benchmark for studying nonlinear dynamics, chaos, and reduced-order modeling in dynamical systems. The training dataset is generated from a single long-term simulation of the Kuramoto-Sivashinsky equation, spanning \( t = 0 \) to \( t = 105 \), with samples recorded every 0.25 time units. Owing to the ergodic nature of the KS system, this extended trajectory effectively captures a wide range of dynamical behaviors and can be partitioned into multiple shorter sub-trajectories with distinct initial conditions. We used this dataset directly from previous studies \citep{linot2022data,linot2023stabilized,chakraborty2024divide}.
	\begin{figure}
		\centering
		\begin{subfigure}[t]{0.48\linewidth}
			\centering
			\includegraphics[width=\linewidth]{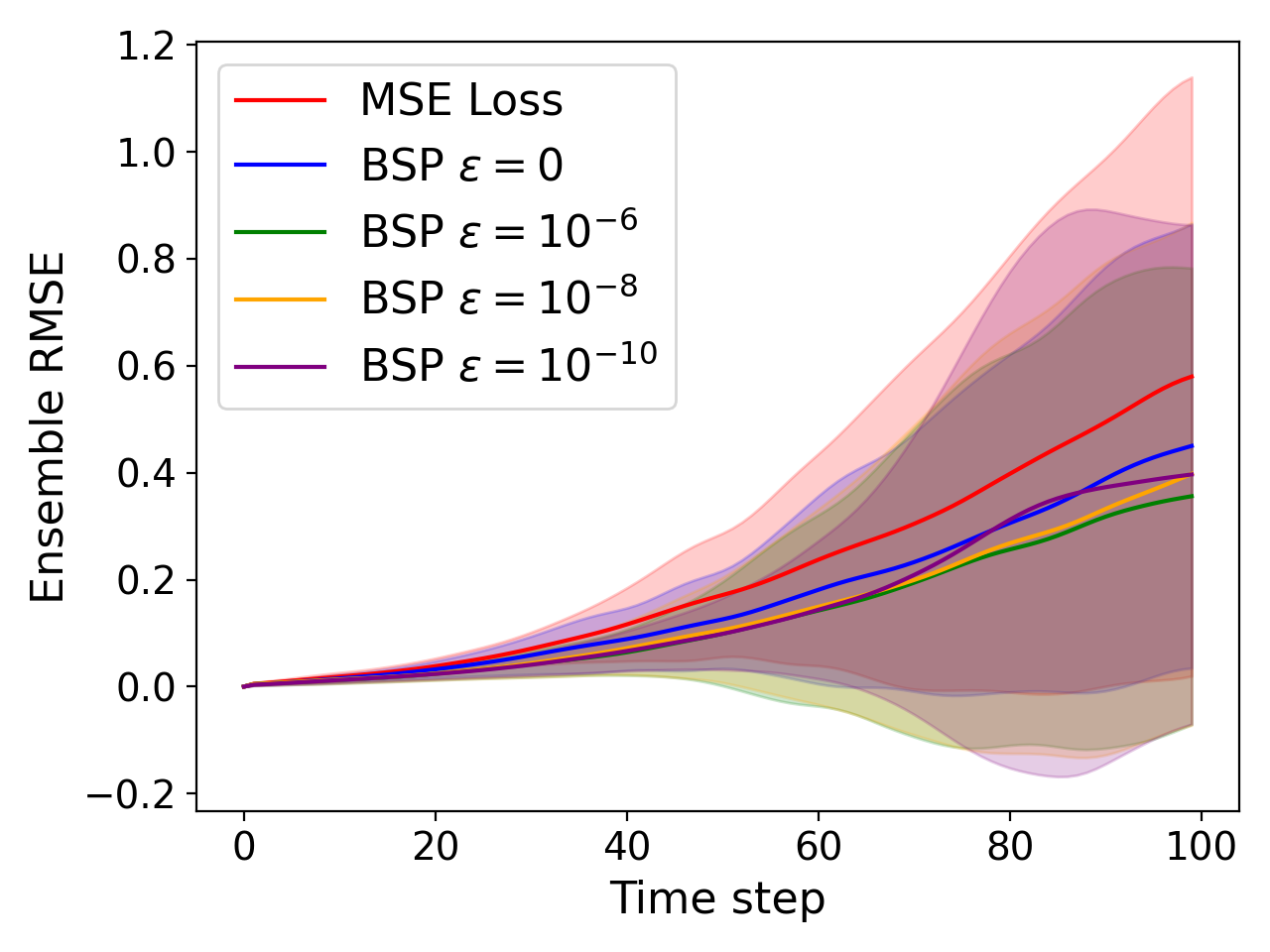}
		\end{subfigure}
		\hfill
		\begin{subfigure}[t]{0.48\linewidth}
			\centering
			\includegraphics[width=\linewidth]{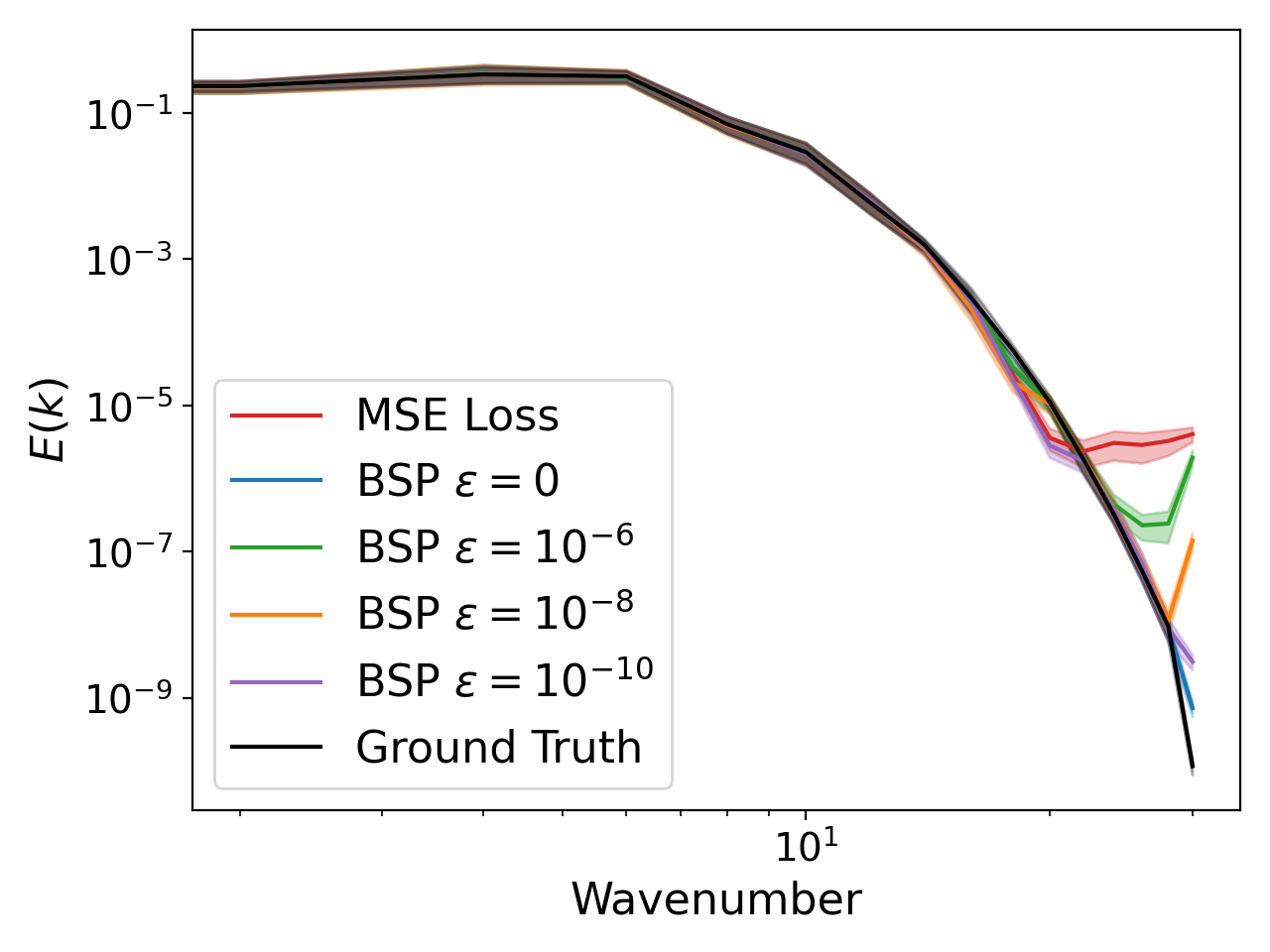}
		\end{subfigure}
		\caption{Comparison of MSE and BSP-trained models across two diagnostics: (a) RMSE : BSP-trained models achieve consistently lower RMSE than MSE. Larger values of $\varepsilon$ show better RMSE. (b) Energy spectrum $E(k)$. BSP loss improves spectral fidelity, particularly for smaller values of $\varepsilon$ (e.g., $0, 10^{-10}$). Shaded regions denote $1\sigma$ ensemble variability.}
		\label{fig:KS_rmse_spec}
	\end{figure}
	
	We implement a recurrent forecasting model using a two-layer Long Short-Term Memory (LSTM) network \cite{hochreiter1997long}. The model processes input sequences of dimension 64 and projects the final hidden state of the 2-layer LSTM (with 128 hidden units) through a fully connected layer to produce a 64-dimensional output. The LSTM captures temporal dependencies in the input sequence, enabling the model to learn effective representations for time series prediction. Forecasting is done in an autoregressive manner. We chose this model to show the ability of BSP loss to work with different model architectures. We observe that in this case, the baseline model itself remains stable in long-term surrogate modeling.  Therefore, we perform an ablation study by implementing the BSP loss with values of $\varepsilon \in \{0, 10^{-6}, 10^{-8}, 10^{-10}\}$ and compare it with the model trained with just the MSE loss.  As shown in \linkref[Fig.]{\ref{fig:KS_image}}, models trained with BSP loss exhibit consistently lower ensemble RMSE over time, with larger $\epsilon$ values yielding improved medium-range forecasting accuracy. Spectral analysis further confirms that BSP loss trained with any $\epsilon$ value aligns spatial structures at different scales more closely with ground truth (refer to \linkref[Fig.]{\ref{fig:KS_rmse_spec}}b). The tradeoff between better medium-range forecast and better spatial structure fidelity for high and low $\varepsilon$ respectively, can be clearly seen from \linkref[Table]{\ref{tab:KS_table}}.
	
	\begin{table}
		\tbl{Comparison of total RMSE over timesteps (0 to 100) and relative spectrum RMSE for models trained with MSE loss and BSP loss at varying $\varepsilon$. The lowest error in each column is highlighted in \textbf{bold}. The relative RMSE is chosen for the energy spectrum due to varying scales.}{%
			\begin{tabular}{lll}
				\toprule
				{Model} & {Forecast RMSE} & $\boldsymbol{E(k)}${ relative RMSE} \\
				\colrule
				MSE Loss & 0.2112 ± 0.1747 & 2283.2818 ± 696.5619 \\
				BSP Loss ($\epsilon=10^{-6}$) & \textbf{0.1313 ± 0.1114} & 1081.0023 ± 348.6294 \\
				BSP Loss ($\epsilon=10^{-8}$) & 0.1385 ± 0.1180 & 79.1884 ± 26.0279 \\
				BSP Loss ($\epsilon=10^{-10}$) & 0.1459 ± 0.1320 & 1.6356 ± 0.5601 \\
				BSP Loss ($\epsilon=0$) & 0.1632 ± 0.1352 & \textbf{0.3638 ± 0.2560} \\
				\botrule
		\end{tabular}}
		\label{tab:KS_table}
		
	\end{table}
	
	\subsection{Two-dimensional turbulence}\label{Xsec11-4.3}\label{2D Turb Section}
	
	Forced two-dimensional turbulence is a standard benchmark for dynamical system prediction due to its chaotic behavior~\citep{stachenfeld2021learned,schiff2024dyslim,frerix2021variational}. We evaluate our proposed loss on 2D homogeneous isotropic turbulence with Kolmogorov forcing, governed by the incompressible Navier-Stokes equations. The two-dimensional Navier-Stokes equations are given by:
	
	\begin{equation}
		\begin{aligned}
			\frac{\partial \mathbf{u}}{\partial t} + \nabla \cdot (\mathbf{u} \otimes \mathbf{u}) &= \frac{1}{Re} \nabla^2 \mathbf{u} - \frac{1}{\rho} \nabla p + \mathbf{f}, \\
			\nabla \cdot \mathbf{u} &= 0,
		\end{aligned}
		\label{Xeqn14-14}
	\end{equation}
	
	where $\mathbf{u} = (u, v)$ is the velocity vector, $p$ is the pressure, $\rho$ is the density, $Re$ is the Reynolds number, and $\mathbf{f}$ represents the forcing function, defined as:
	\begin{figure}
		\centering
		\begin{minipage}[t]{0.53\linewidth}
			\vspace{-.1cm}
			\hspace{-0.2cm}
			\includegraphics[width=1.05\linewidth]{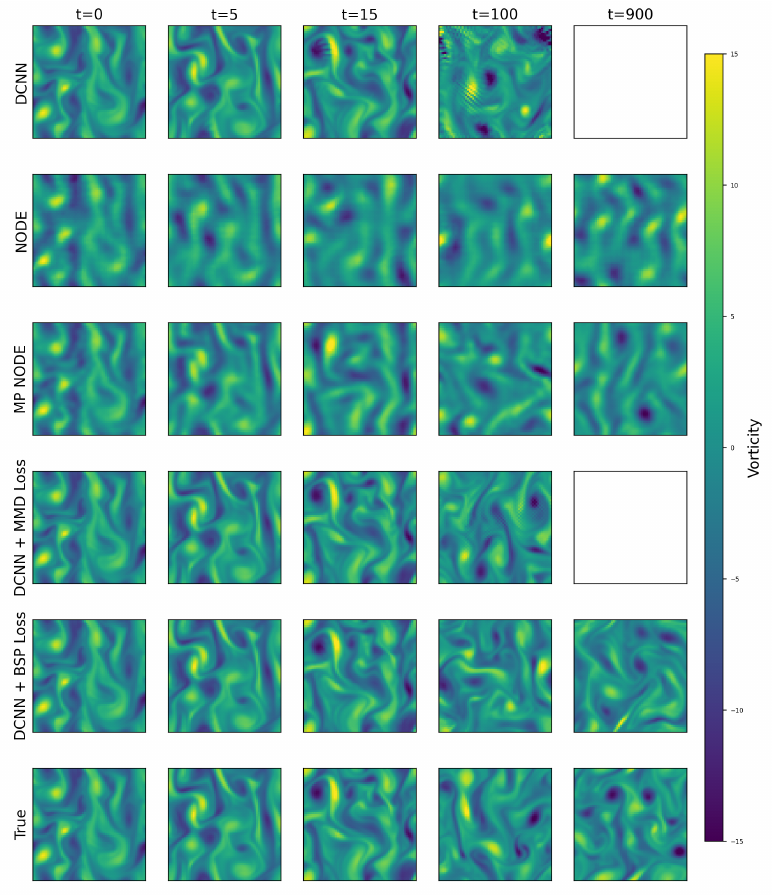}
		\end{minipage}
		\hfill
		\begin{minipage}[t]{0.4\linewidth}
			\vspace{-0.4cm}
			\includegraphics[width=1.1\linewidth]{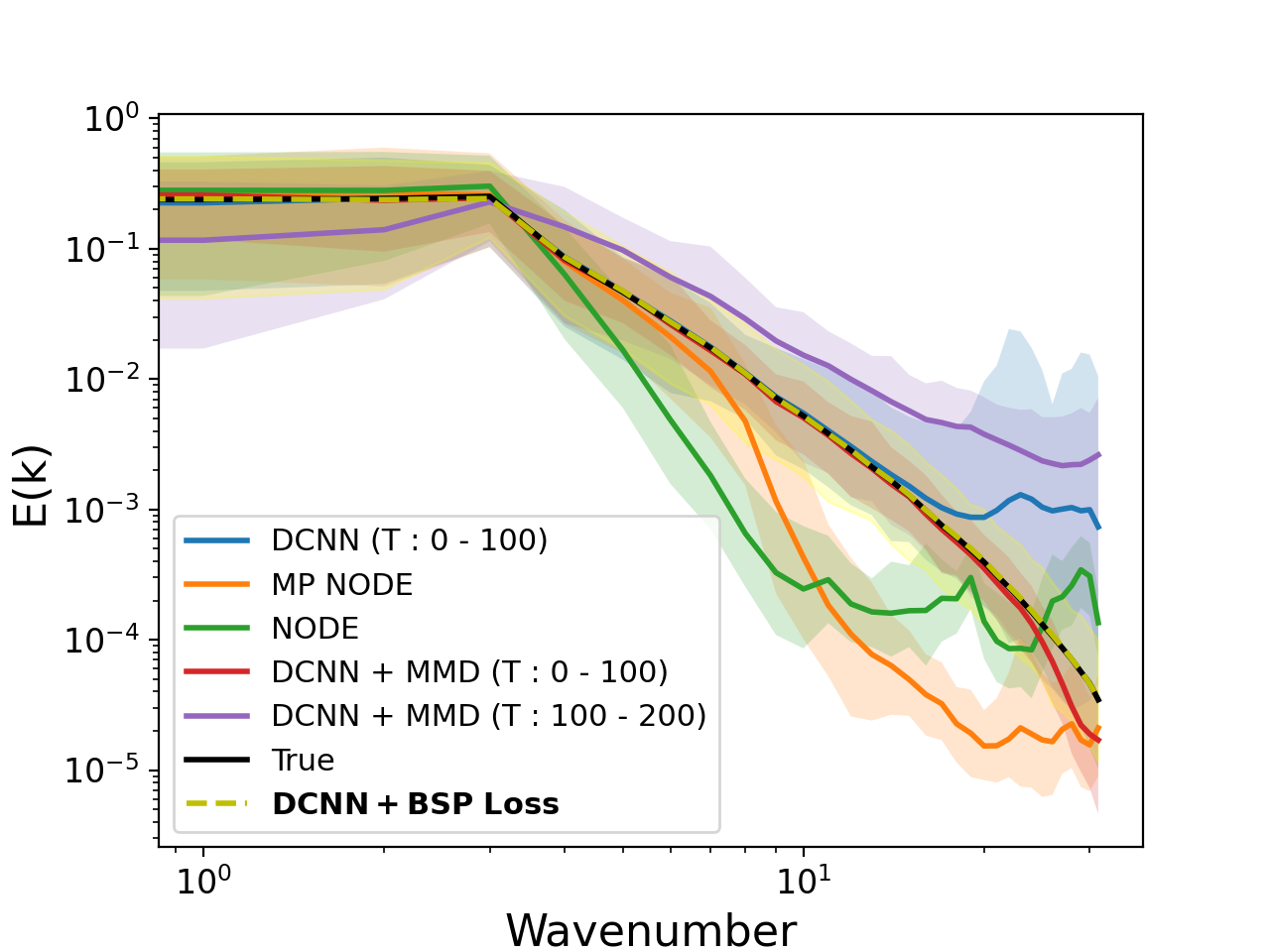}
			\vspace{0.2cm}
			\includegraphics[width=1.1\linewidth]{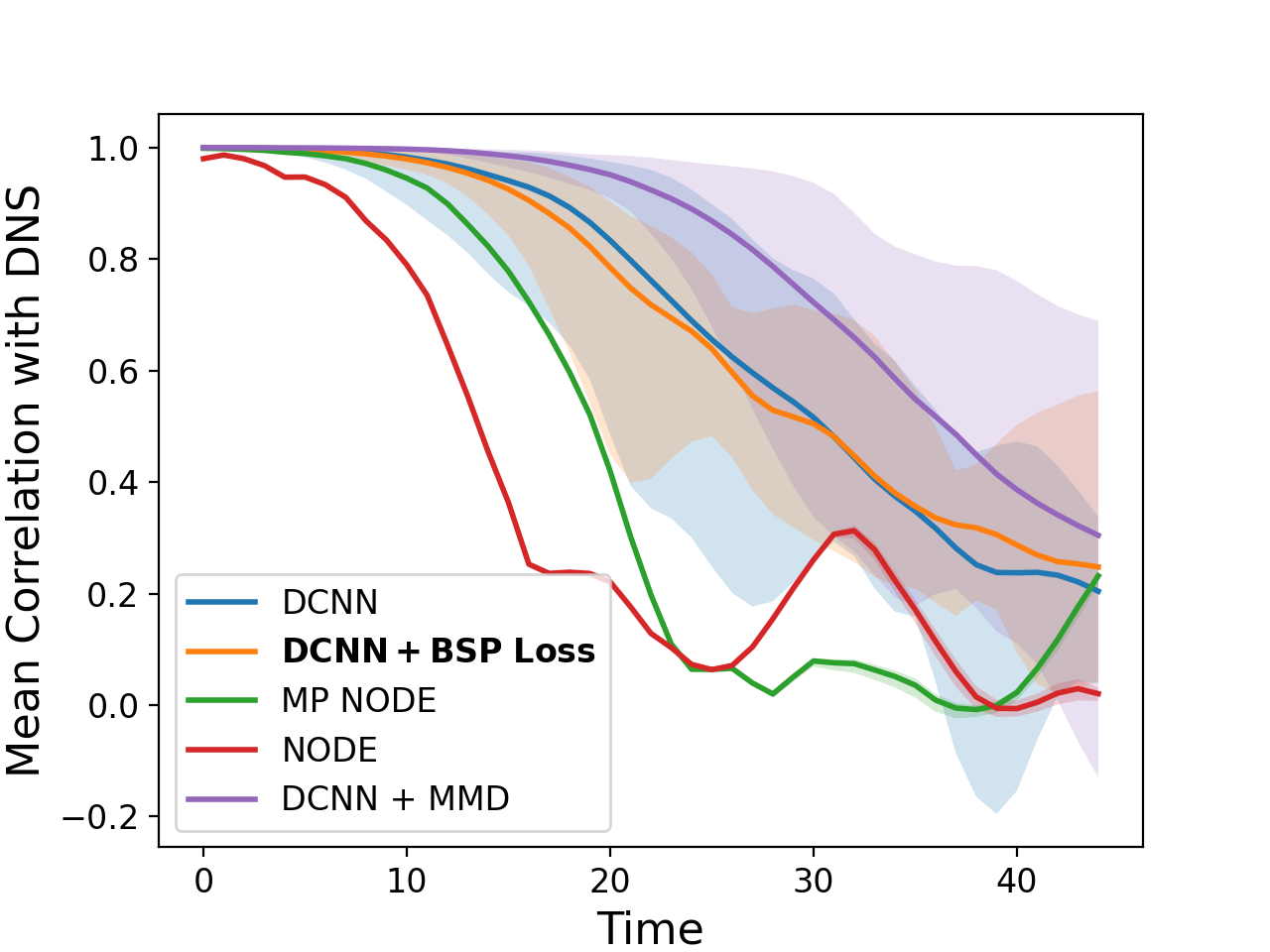}
		\end{minipage}
		\caption{Comparison of NODE, MP-NODE, and DCNN models with MSE, MMD, and BSP losses. (a) shows spatial accuracy and stability over time; (b-c) summarizes spectral fidelity and correlation behavior. BSP matches the ground truth energy spectrum best over 900 steps. MMD aligns the best at short times (t<100) but degrades later. Overall, BSP maintains structure and energy distribution across long forecast horizons.}
		\label{fig:2d_all}
	\end{figure}
	\begin{equation}
		\mathbf{f} = A \sin(ky)\hat{\mathbf{e}} - r\mathbf{u},
		\label{Xeqn15-15}
	\end{equation}
	
	with parameters $A = 1$ (amplitude), $k = 4$ (wavenumber), $r = 0.1$ (linear drag), and $Re = 1000$ (Reynolds number) selected for this study as given in \citep{shankar2023differentiable}. Here, $\hat{\mathbf{e}}$ denotes the unit vector in the $x$-direction. The initial condition is a random divergence-free velocity field~\citep{Kochkov2021-ML-CFD}.
	The ground truth datasets are generated using direct numerical simulations (DNS)~\citep{kochkov2021machine} of the governing equations within a doubly periodic square domain of size $L = 2\pi$, discretized on a uniform $512 \times 512$ grid and filtered to a coarser $64 \times 64$ grid. The trajectories are sampled temporally after the flow reaches the chaotic regime, with snapshots spaced by $T = 256\Delta t_{DNS}$, ensuring sufficient distinction between consecutive states. Details of the dataset construction can be found in the work by \citep{shankar2023differentiable}.
	
	\begin{figure}
		\centering
		\begin{subfigure}[b]{0.45\textwidth}
			\includegraphics[width=\textwidth]{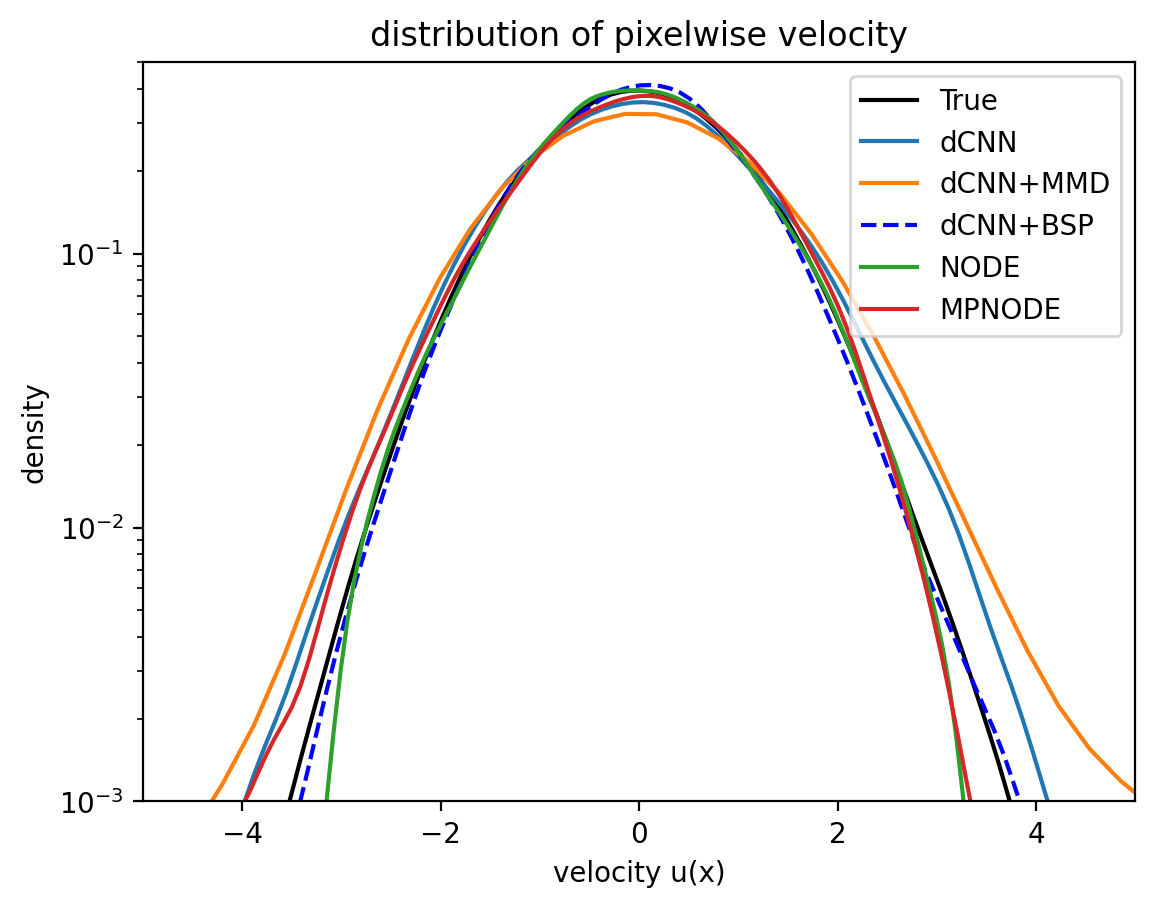}
		\end{subfigure}
		\begin{subfigure}[b]{0.45\textwidth}
			\includegraphics[width=\textwidth]{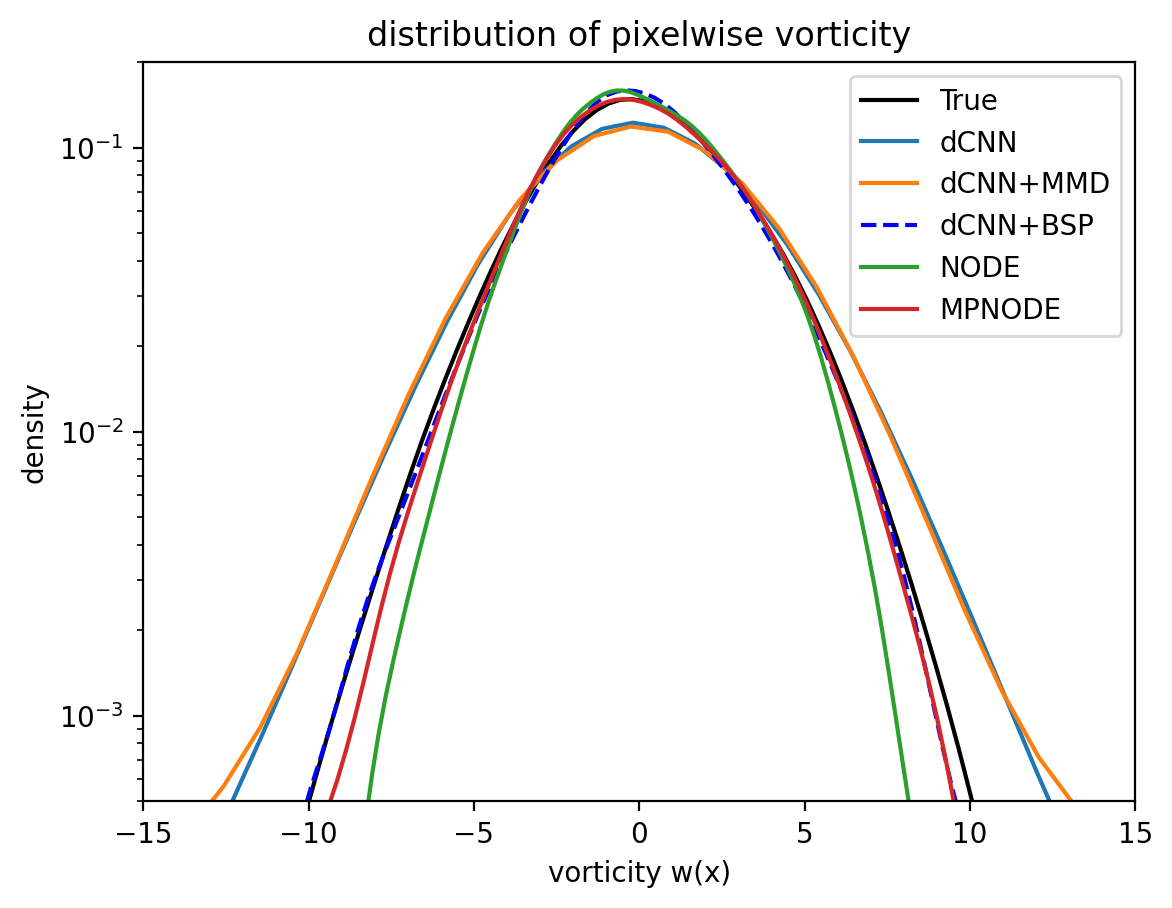}
		\end{subfigure}

		\begin{subfigure}[b]{0.45\textwidth}
			\includegraphics[width=\textwidth]{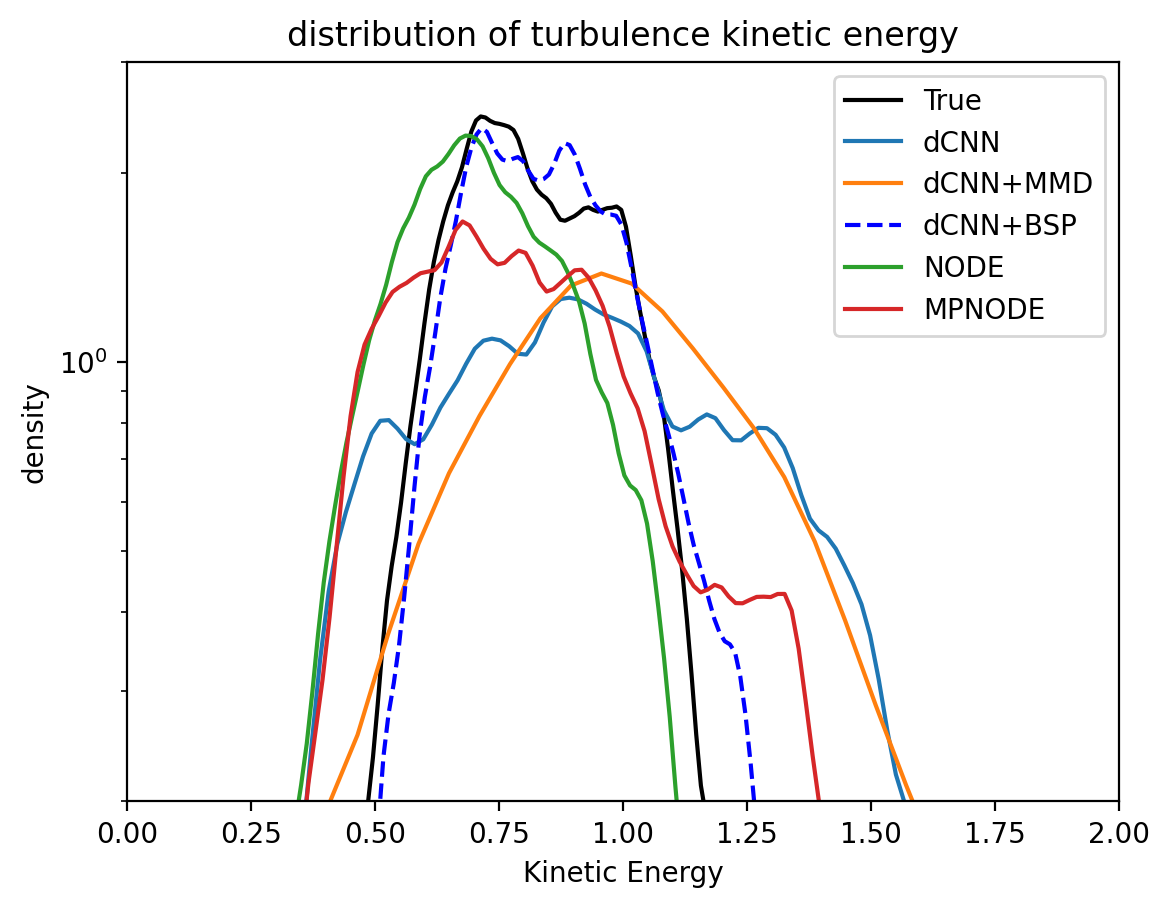}
		\end{subfigure}
		\begin{subfigure}[b]{0.45\textwidth}
			\includegraphics[width=\textwidth]{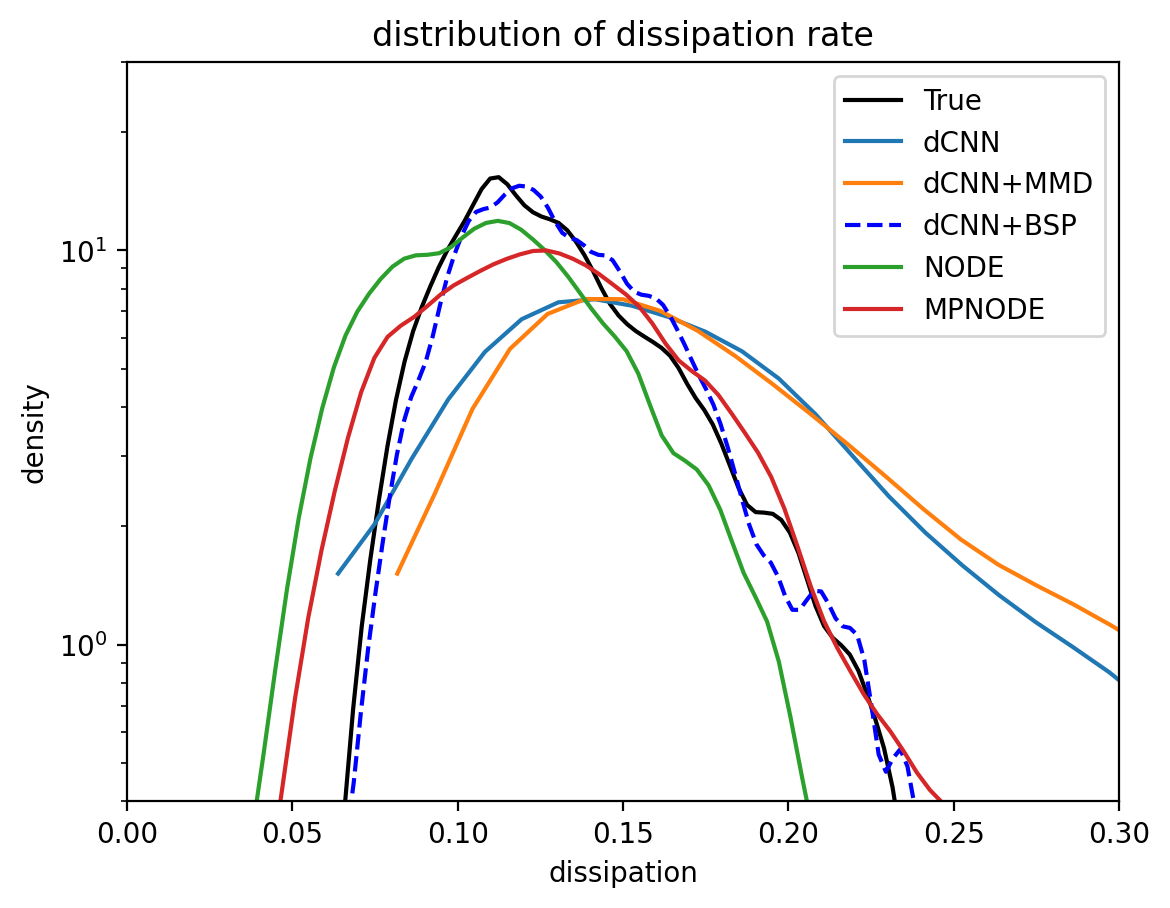}
		\end{subfigure}
		\caption{Comparison of the probability density functions (PDFs) of various invariant physical quantities of different model predictions against the true data. The quantities shown are distributions of: (top left) pixelwise velocity $u(x)$, (top right) pixelwise vorticity, (bottom left) turbulence kinetic energy, and (bottom right) dissipation rate. The models compared include a baseline deterministic convolutional neural network (dCNN), dCNN with maximum mean discrepancy (MMD) loss, dCNN with Binned Spectral Power (BSP) loss, a Neural Ordinary Differential Equation (NODE), and  a Multi-step Penalty NODE (MPNODE). The distribution of quantities for models trained with BSP loss (dashed blue line) is the closest to the ground truth(solid black line) for all the invariant quantities.}
		\label{fig:invariant_KF}
	\end{figure}
	
	All baseline models are trained using the multi-step rollout loss from \linkref[\del{Equation}\ins{Eq.}]{\ref{MSE-2}} and the \textit{pushforward-trick}. We use the dilated Convolutional Neural Network (DCNN) architecture~\citep{stachenfeld2021learned}, with hyperparameters listed in \ref{hyperparams}. For this test case as well as the following example in \linkref[Section]{\ref{3d_turb}}, we use $ \lambda_i$ as $k_{(bin\ i)}^2$ following widely used procedures in literature \citep{shankar2023differentiable,oommen2024integrating,li2021markov}. As benchmarks, we include DCNN with Maximum Mean Discrepancy (DCNN + MMD)~\citep{schiff2024dyslim}, which promotes attractor learning for stability, and Neural ODE (NODE) and MP-NODE~\citep{chen2018neural,chakraborty2024divide}, with results taken from \citep{chakraborty2024divide}. \ref{baselines} details these baselines.
	
	\begin{figure}
		\centerline{\includegraphics[width=0.75\linewidth]{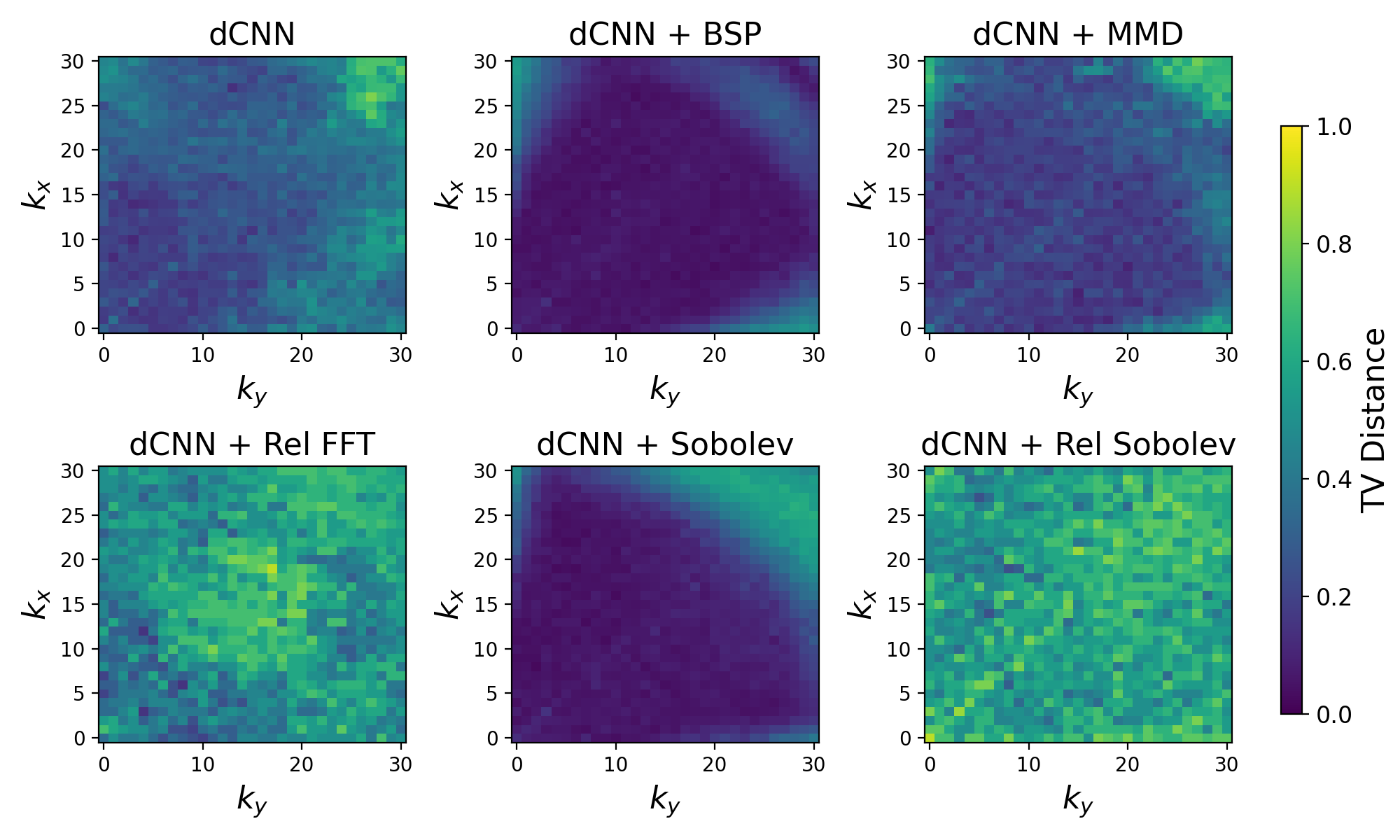}}
		\caption{Total variation (TV) distance between the predicted and true spectral component distributions across wavenumbers $k_x$ and $k_y$ for different loss functions. Among all methods, the model trained with the BSP loss exhibits the lowest TV distance, indicating the closest match to the true spectral distribution and the most effective mitigation of spectral bias.}
		\label{fig:TV distance}
	\end{figure}
	
	\linkref[\del{Figure}\ins{Fig.}]{\ref{fig:2d_all}}a shows that DCNN trained with MSE becomes unstable at longer rollouts, consistent with prior works. DCNN + MMD improves stability up to $t=100$ but becomes unstable after that, diverging in high-wavenumber energy (\linkref[\del{Figure}\ins{Fig.}]{\ref{fig:2d_all}}b) due to failure to capture finer details~\citep{maulik2019subgrid}. NODE and MP-NODE remain stable but fail to preserve small-scale structures. In contrast, DCNN + BSP maintains stability and resolves both large- and small-scale features across the trajectory, preserving the energy spectrum throughout (\linkref[\del{Figure}\ins{Fig.}]{\ref{fig:2d_all}}b). Unlike MMD, the BSP loss does not minimize error in physical space, leading to no significant improvement in correlation metrics here(\linkref[\del{Figure}\ins{Fig.}]{\ref{fig:2d_all}}c). However, for stochastic systems like turbulence, invariant metrics are more meaningful. \linkref[\del{Figure}\ins{Fig.}]{\ref{fig:invariant_KF}} presents a detailed comparison of how well different models reproduce key physical invariants of the underlying dynamics by plotting the probability density functions (PDFs) of four important quantities: pixel-wise $u(x)$ velocity, vorticity, turbulence kinetic energy (TKE), and dissipation rate. These metrics are crucial because they characterize both large-scale flow structures and small-scale turbulent behaviors, providing a comprehensive assessment of the physical fidelity of the models. The models evaluated include the same baselines as mentioned in \linkref[\del{section}\ins{Section}]{\ref{2D Turb Section}}. The results show that across all four quantities, the model trained with BSP loss (shown by a dashed blue line) produces distributions that align most closely with the ground truth data (solid black line). This indicates that the BSP loss not only improves spectral accuracy but also enables the model to better capture the complex statistical properties of the underlying dynamical system, outperforming both baseline loss functions and other models like NODE and MPNODE in preserving invariant physical characteristics.
	
	We also benchmark against other spectral losses: Sobolev~\citep{li2021markov}, relative FFT, and relative Sobolev. The total variation (TV) distance~\citep{wang2024beyond} is employed to quantify discrepancies between the spectral component distributions at different wavenumbers, providing a robust measure of how predicted and true spectra differ across scales. As shown in \linkref[\del{Figure}\ins{Fig.}]{\ref{fig:TV distance}}, BSP outperforms other losses in spectral fidelity. We note that the Sobolev loss also shows decent performance. We hypothesize that the poor performance of the relative losses is due to them trying to minimize very small values in the Fourier domain in a point-to-point manner, which is nontrivial. This justifies our use of binning to capture the energy at different scales in the BSP loss.
	
	\subsection{3D Turbulence}\label{Xsec12-4.4}
	
	\label{3d_turb}
	\begin{figure*}
		
		\centerline{\includegraphics[width=\textwidth]{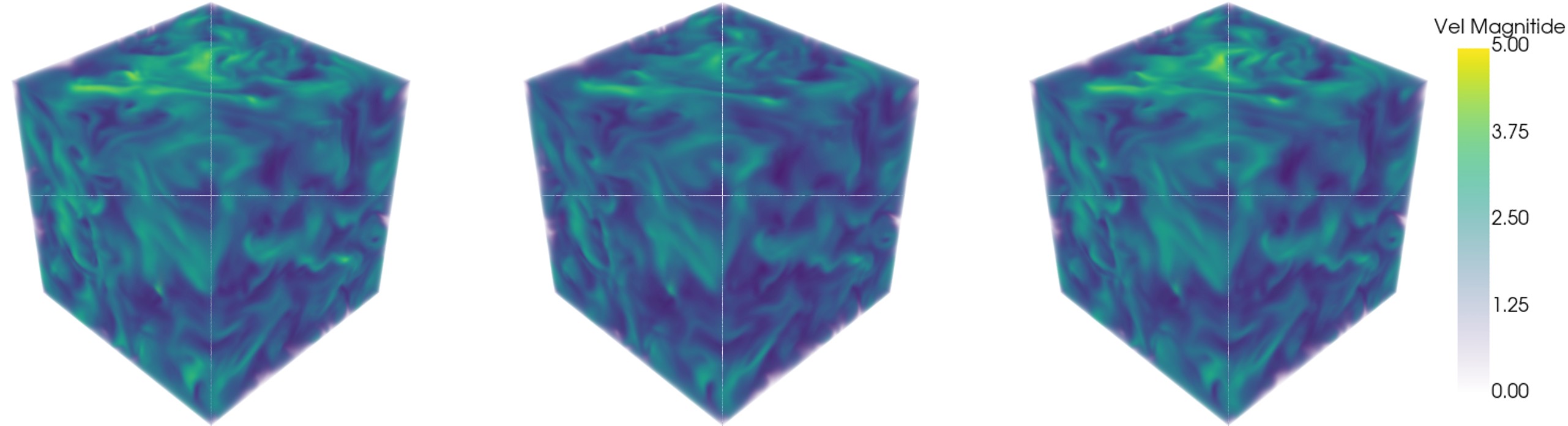}}
		\caption{Velocity magnitude 3D plot for ground truth(left), UNet prediction(mid), and UNet + BSP loss prediction(right) after 5 auto-regressive rollouts. Clearly, the UNet prediction has some blurring effect that is rectified by the BSP loss.}
		\label{fig:3D_turb_diagram}
	\end{figure*}
	
	\begin{figure}
		
		\centerline{\includegraphics[width=0.75\columnwidth]{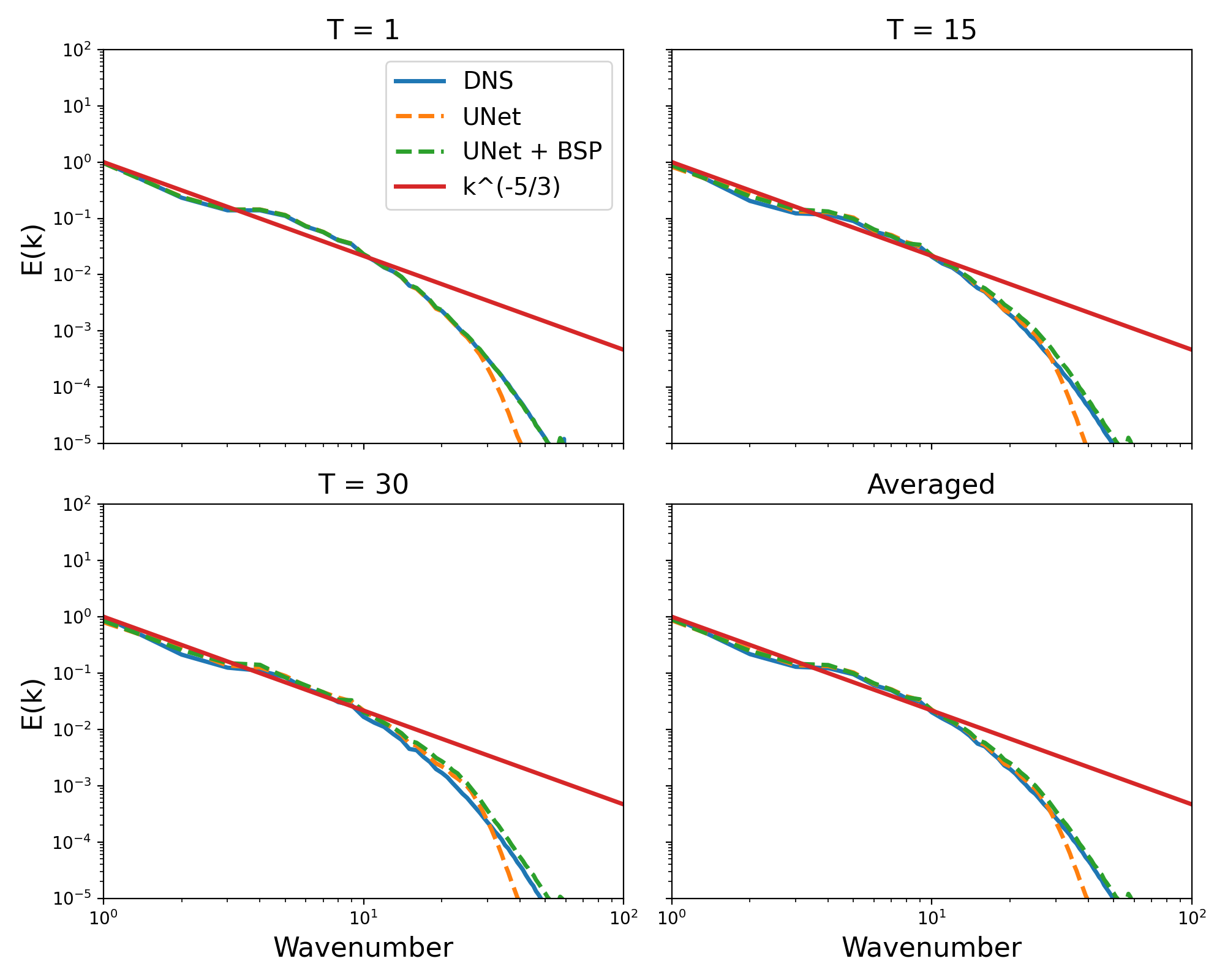}}
		\caption{Comparison of energy spectra $E(k)$ as a function of wavenumber at different time steps ($T = 1, 15, 30$) and averaged over time. The plots show results from DNS (blue solid line), UNet (orange dashed line), and UNet model trained with BSP loss (green dashed line), along with the theoretical $k^{-5/3}$ scaling \citep{kolmogorov1941local} (red solid line). The inclusion of BSP improves the spectral accuracy at high wavenumbers compared to the standalone UNet approach.}
		\label{fig:3dspec}
	\end{figure}
	This experiment uses data from a three-dimensional direct numerical simulation (DNS) of incompressible, homogeneous, isotropic turbulence \citep{mohan2020spatio}. The computational domain is a cubic box with dimensions of $128^3$ grid points. Two scalar fields, each with distinct probability density function (PDF) characteristics, are advected as passive scalars by the turbulent flow. This dataset is taken from \citep{mohan2020spatio}. They refer to this dataset as \textit{ScalarHIT}, following \citep{daniel2018reaction}. The DNS is performed with a pseudo-spectral code, ensuring incompressibility via
	\begin{equation}
		\partial_{x_i} v_i = 0,
		\label{Xeqn16-16}
	\end{equation}
	and solving the Navier--Stokes equations
	\begin{equation}
		\partial_t v_i + v_j \partial_{x_j} v_i = -\frac{1}{\rho}\partial_{x_i} p + \nu \nabla^2 v_i + f^v_i.
		\label{Xeqn17-17}
	\end{equation}
	\begin{figure}
		
		\centerline{\includegraphics[width=0.89\linewidth]{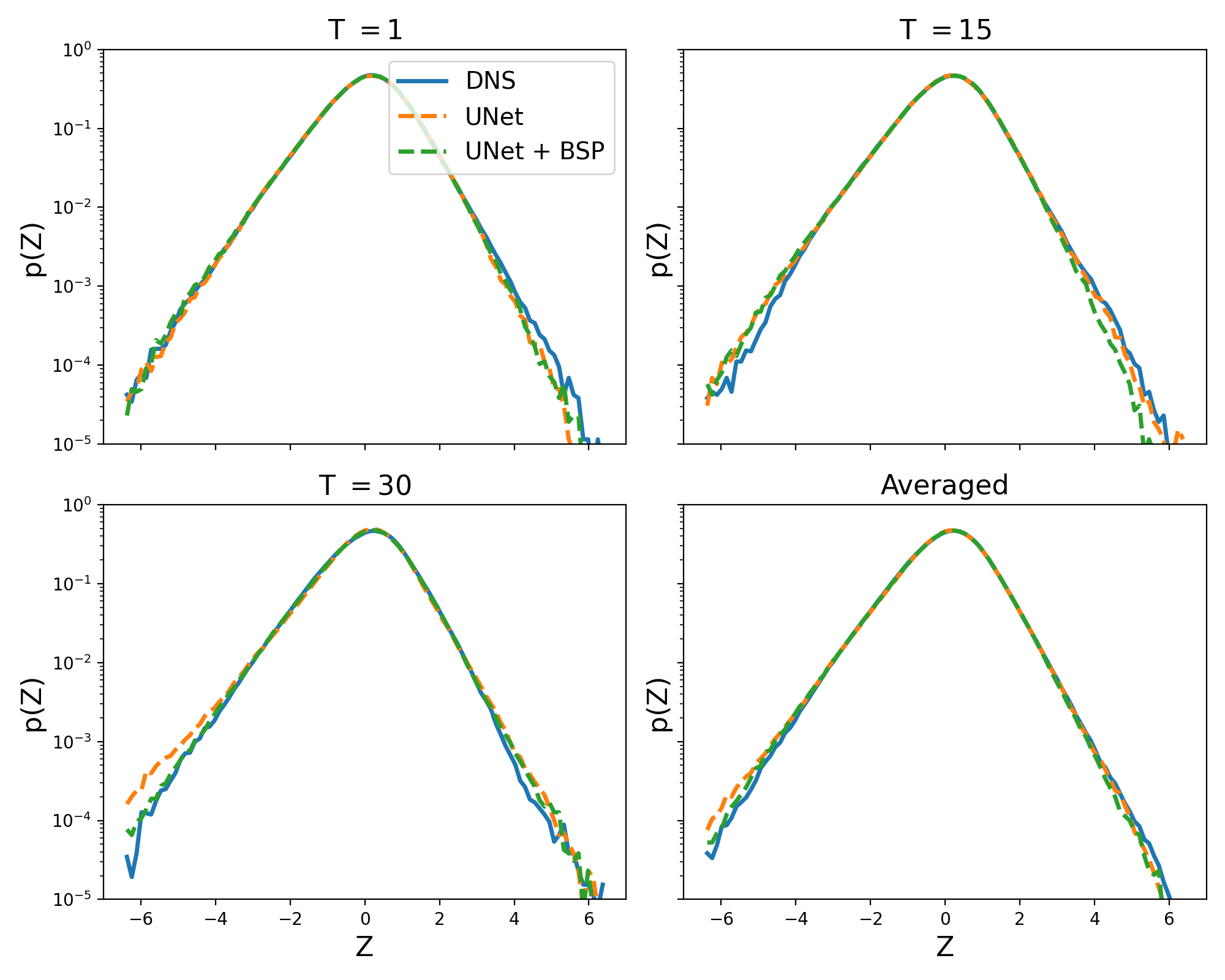}}
		\caption{The figure illustrates the comparison of the intermittency plots for UNet models trained with MSE loss (orange) and UNet trained with BSP loss (green) across different time steps (T). }
		\label{fig:3dintermittency}
	\end{figure}
	Low-wavenumber forcing ($k<1.5$) maintains a statistically steady state. Dealiasing is performed through phase-shifting and truncation, achieving a resolved maximum wavenumber of $k_{\max} \approx 60$ with spectral resolution $\eta k_{\max}\approx 1.5$. Scalar transport is governed by
	\begin{equation}
		\partial_t \phi + v_j \partial_{x_j}\phi = D \nabla^2 \phi + f^\phi,
		\label{Xeqn18-18}
	\end{equation}
	where $\phi$ is a passive scalar and $D$ is its diffusivity. Both the viscosity $\nu$ and diffusivity $D$ are chosen so that the Schmidt number $Sc=\nu/D=1$. The integral-scale Reynolds number is expressed in terms of the Taylor microscale as
	\begin{equation}
		Re_\lambda = \sqrt{\frac{20}{3}}\frac{\text{TKE}}{\nu},
		\label{Xeqn19-19}
	\end{equation}
	\begin{figure}
		
		\centerline{\includegraphics[width=0.99\linewidth]{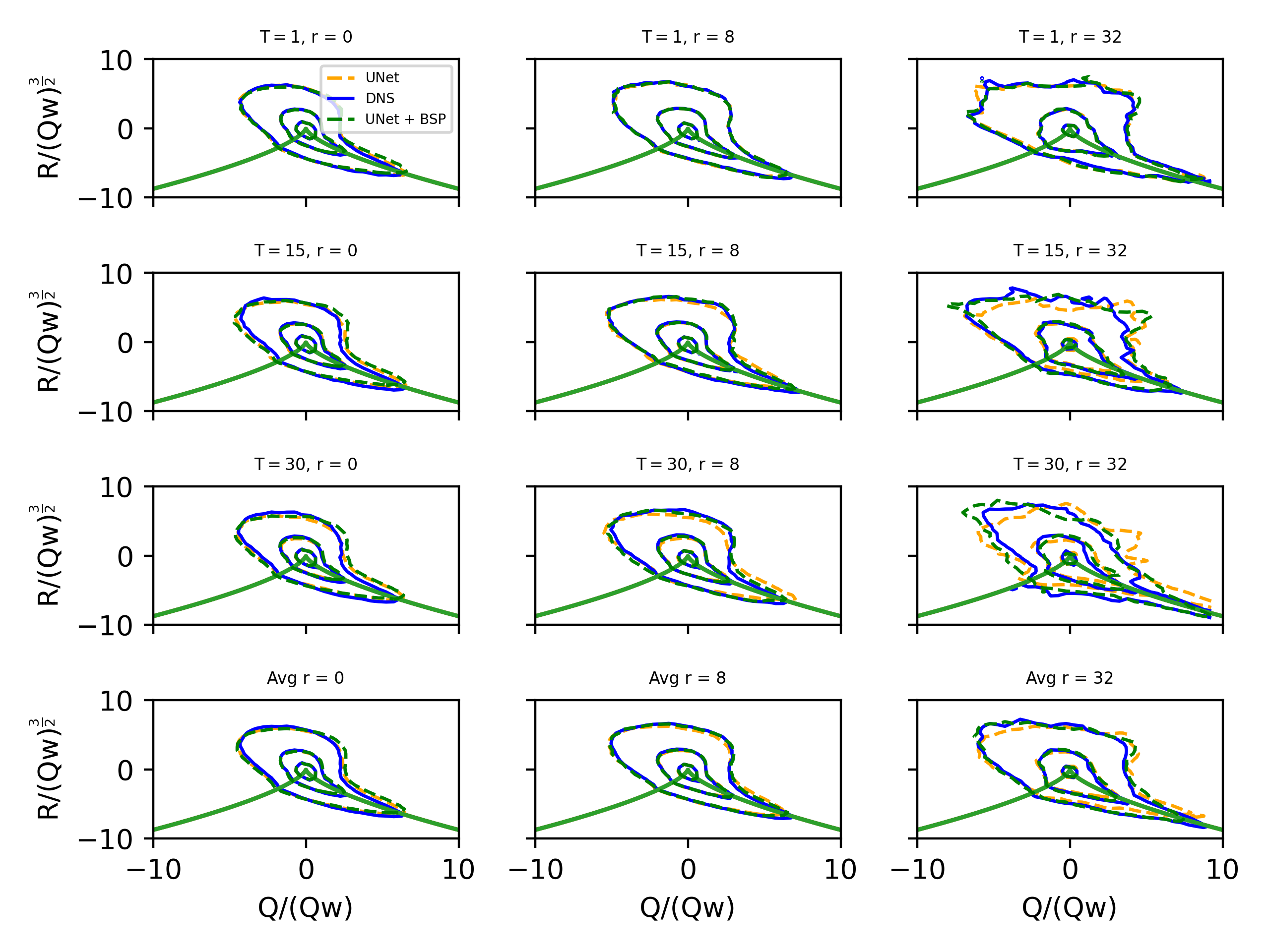}}
		\caption{The figure illustrates the comparison of the QR plots for UNet models trained with MSE loss (orange) and UNet trained with BSP loss (green) across different time steps (T) and resolutions (r). The QR plots signify three dimensional chaos in turbulence.}
		\label{fig:3dqr}
	\end{figure}
	where TKE denotes the turbulent kinetic energy. They use a novel scalar forcing approach, inspired by chemical reaction kinetics \citep{daniel2018reaction} to achieve desired stationary scalar PDFs and ensure scalar boundedness. Assuming scalar bounds $\phi_l=-1$ and $\phi_u=+1$, the forcing term is modeled as
	\begin{equation}
		f^\phi = \operatorname{sign}(\phi) f_c |\phi|^n (1 - |\phi|)^m,
		\label{Xeqn20-20}
	\end{equation}
	where $f_c$, $m$, and $n$ adjust PDF shape and scalar distribution. By appropriate parameter choices, different scalar PDFs are realized. For the present dataset, one scalar exhibits near-Gaussian behavior (kurtosis $\approx 3$) while the other has a lower kurtosis ($\approx 2.2$). With this forcing, the velocity and scalar fields reach a statistically stationary state at $Re_\lambda \approx 91$. Two scalars with distinct PDFs allow for testing model capabilities to reproduce both Gaussian-like and bounded scalar distributions.
	We use a UNet based architecture for both MSE and BSP loss implementation. The hyperparameters of the model are mentioned in the \linkref[Appendix]{\ref{hyperparams}}. In \linkref[\del{Figure}\ins{Fig.}]{\ref{fig:3D_turb_diagram}}, we observe that both the models show minimal spectral bias and improved stability. This is related to the reduced spectral bias of models with larger parameter space(refer Appendix A.5. in \citep{rahaman2019spectral}). We limit the extent of forecasting in this experiment due to limited training and validation data. We tested all models with 30 autoregressive rollouts, which represents approximately one cycle of turbulence for this dataset. \linkref[\del{Figure}\ins{Fig.}]{\ref{fig:3D_turb_diagram}} shows the model trained with BSP loss captures the fine scales better visually. It is also evident from \linkref[\del{Figure}\ins{Fig.}]{\ref{fig:3dspec}} that the BSP loss shows a marked accuracy in the energy spectrum at high wavenumbers, corresponding to dynamically important small-scale structures in chaotic systems. With this evidence, we can conclude that the BSP loss helps in preserving the distribution of energy across different scales and spatial structures.
	
	In \linkref[Fig.]{\ref{fig:3dintermittency}}, we present the intermittency plots. Intermittency refers to the fluctuations in velocity gradients, leading to deviations from Gaussian statistics. This can be analyzed using the probability density function (PDF) of the velocity gradient tensor, which often exhibits heavy tails due to strong localized fluctuations and is a harder quantity to learn correctly \citep{mohan2020spatio}. The tensor, defined as the spatial derivatives of the velocity components, captures small-scale structures where intermittency effects are most pronounced. We observe near-perfect prediction at high frequencies, represented by the tails of the PDF.
	
	Finally, the most stringent test of this method is presented in the Q-R plane spectra in \linkref[Fig.]{\ref{fig:3dqr}}, which represents the three-dimensional chaos in turbulence. QR plots are used to analyze the local flow topology by examining the invariants of the velocity gradient tensor~\citep{chertkov1999lagrangian}. The second invariant, Q, represents the balance between rotational and strain effects, while the third invariant, R, characterizes the nature of vortex stretching and flow structures. The spectra at $r=0$ indicate high frequencies, while those at $r=8$ and $r=32$ indicate intermediate frequencies and low frequencies, respectively. Historically, ML methods have struggled to capture the $r=0$ spectra and instead predict Gaussian-like noise~\citep{mohan2020spatio}, but we show that the BSP loss accurately captures these dynamics without compromising dynamics at $r=8, 32$. These plots show that even after conserving the smaller structures in the flow, the predictions do not deviate from key characteristics of turbulence.
	
	\subsection{Turbulent flow over an airfoil}\label{Xsec13-4.5}\label{airfoil}
	
	In this section, we examine the turbulent wake flow downstream of a NACA0012 airfoil operating at a Reynolds number of 23,000, a free-stream Mach number of 0.3, and an angle of attack of $6^\circ$. We utilize a large eddy simulation (LES) dataset provided by \citep{towne2023database}, available through the publicly accessible \emph{Deep Blue Data} repository from the University of Michigan. The flow features have coherent structures associated with Kelvin-Helmholtz instability over the separation bubble and Von-K\'arm\'an vortex shedding in the wake, while exhibiting features at multiple scales characteristic of turbulent flows. This makes it an ideal test case for several experiments, including validating computational fluid dynamics (CFD) models, analyzing flow dynamics, and exploring reduced-order modeling approaches. For more details on the dataset, the reader is directed to Section VII in \citep{towne2023database}. We follow the same data pre-processing strategy as given in \citep{oommen2024integrating}. The field is interpolated to convert it to a rectangular domain (200x400\del{ }\ins{~}pixels). We implement a UNet architecture \citep{ronneberger2015u} for the base model and improve it by using our BSP loss. The hyperparameters of the model are mentioned in \ref{hyperparams}.
	\begin{figure*}
		
		\centerline{\includegraphics[width=0.99\linewidth,trim=1cm 3cm 0 3cm,clip]{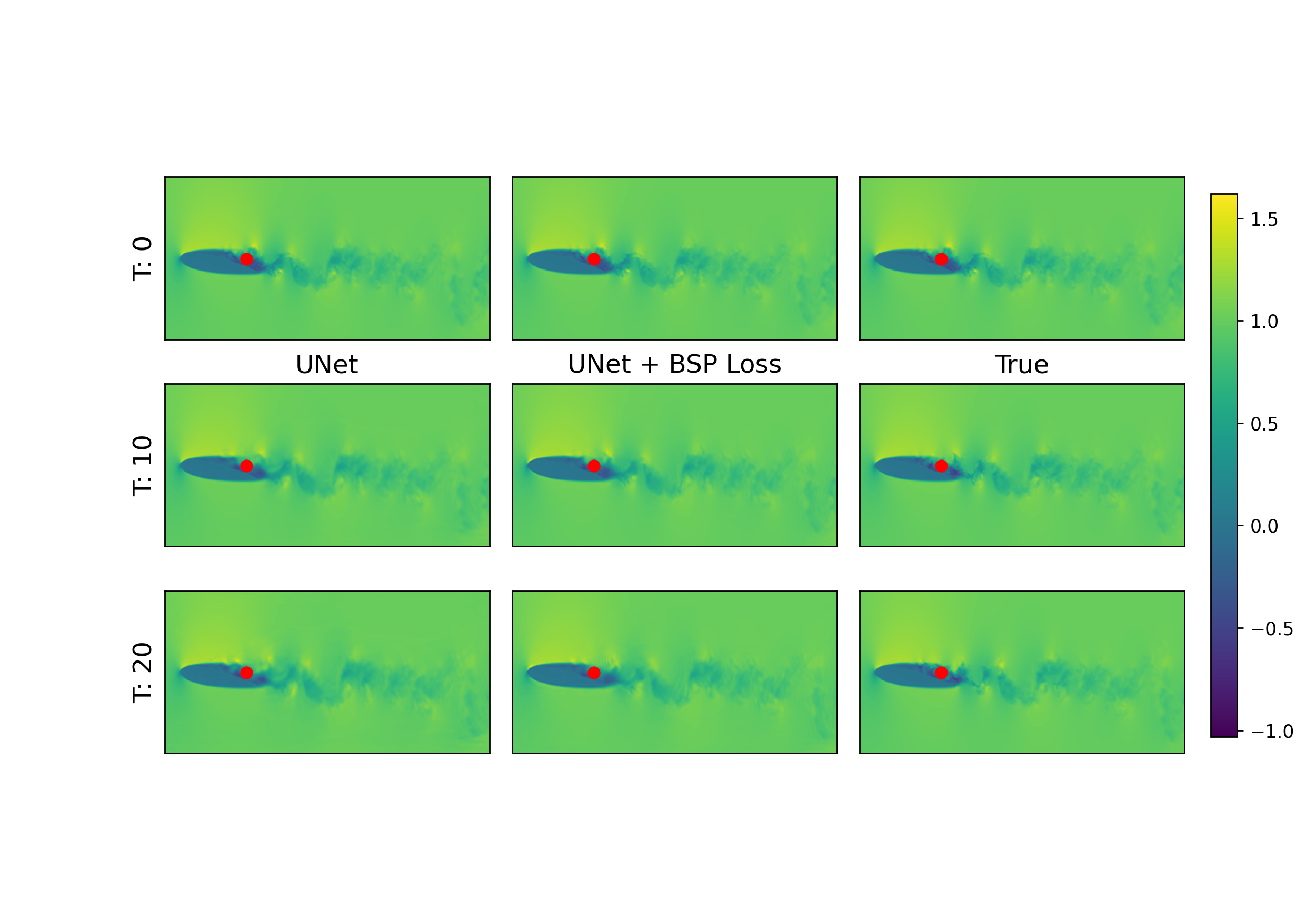}}
		\caption{Comparison of model predictions at different timesteps for UNet (trained with MSE loss) and UNet + BSP Loss. The red dot is the point where the PDF is computed.}
		\label{fig:airfoil}
	\end{figure*}
	\par
\begin{figure}
		
		\includegraphics[width=0.49\linewidth]{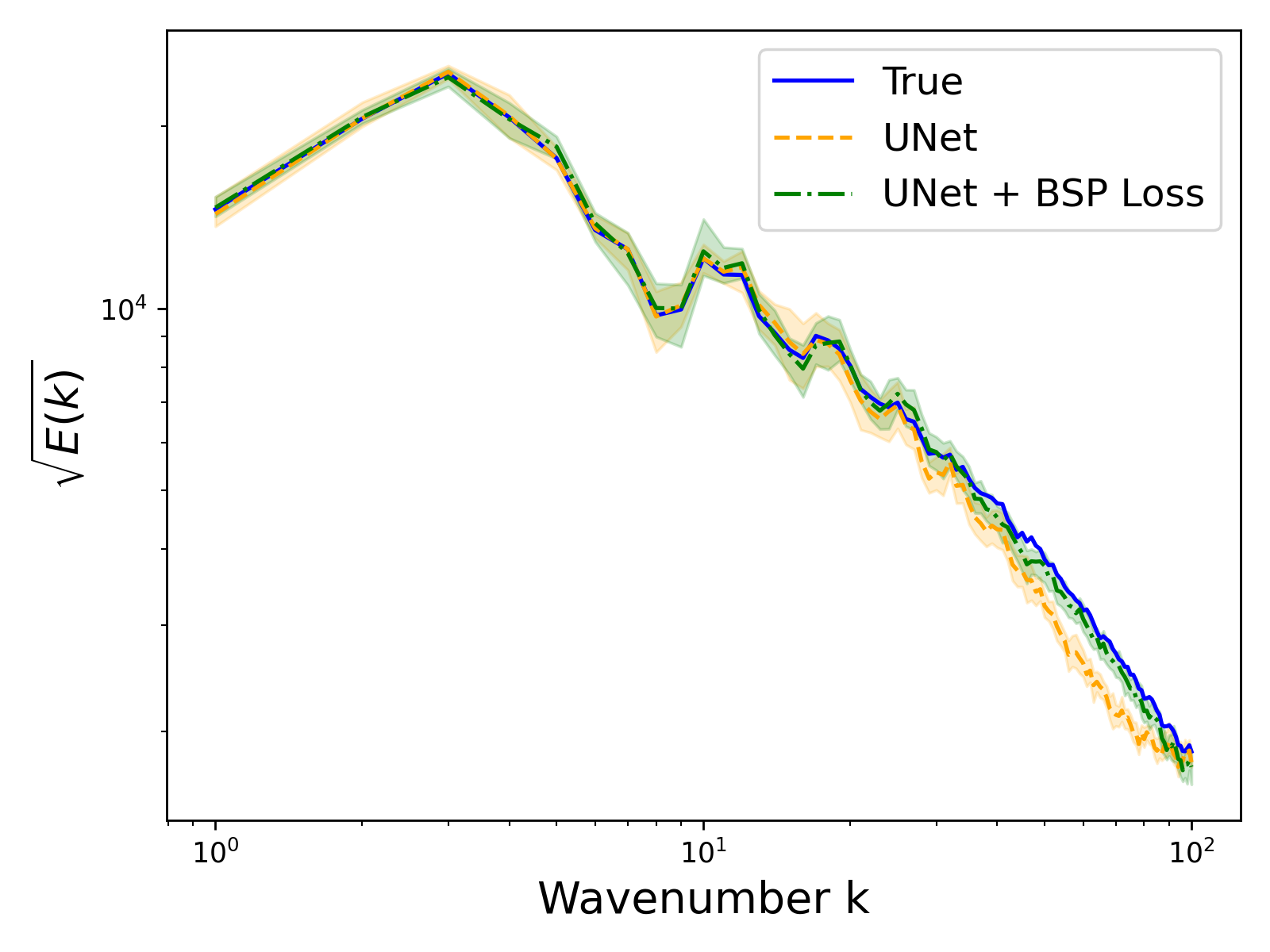}
		\includegraphics[width=0.49\linewidth]{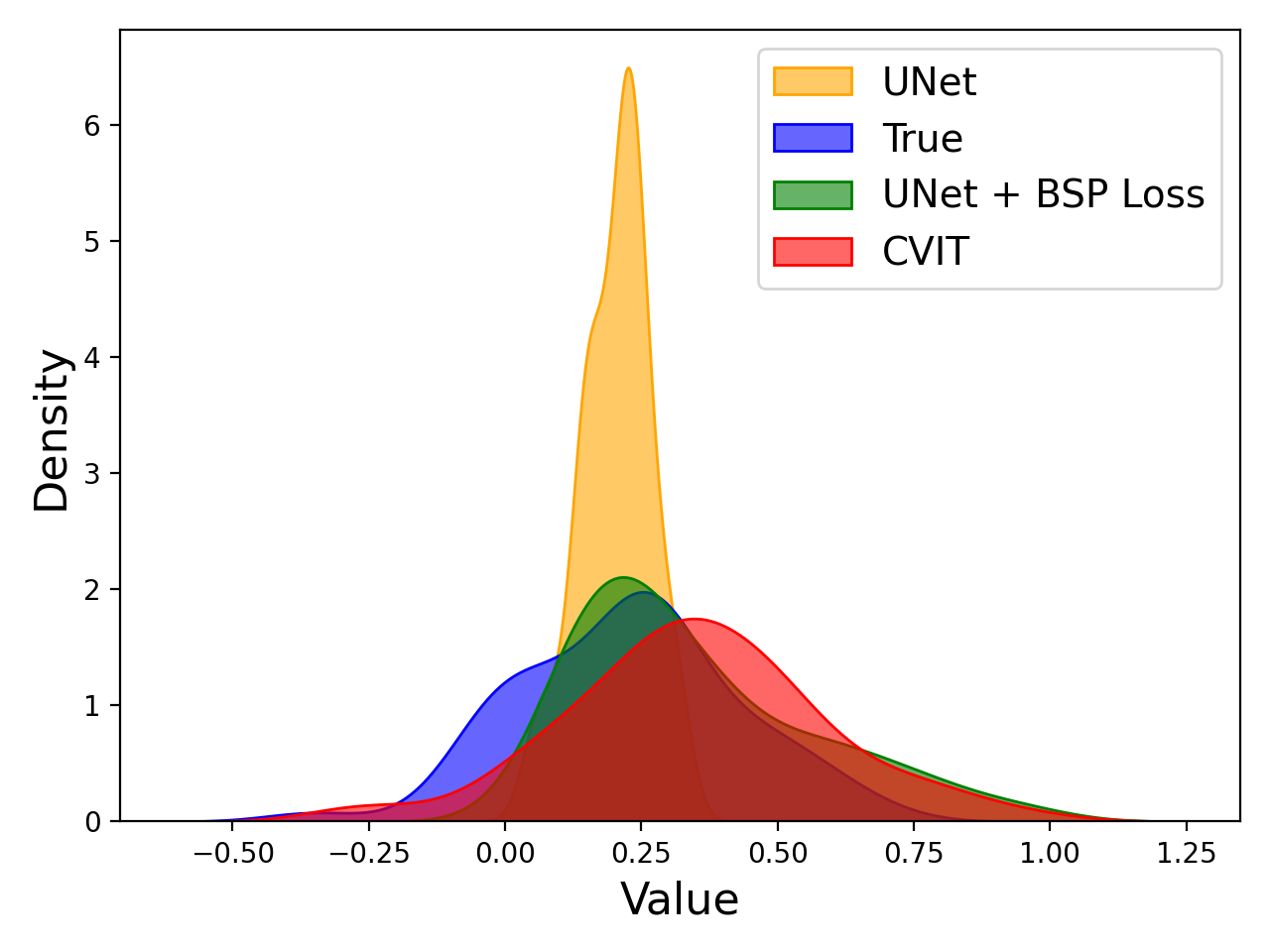}
		\caption{(left)Square root of the energy spectra for ground truth and model predictions. The energy spectra shown here is the mean of the first 10 timestep predictions. (right)Distribution of the velocity field at a location downstream of the airfoil. It shows the comparison of PDFs of ground truth and various model predictions.}
		\label{fig:airfoil-spec}
	\end{figure}

	Contrary to the previous case, here we observed that the energy spectrum of the UNet model prediction is very close to the ground truth even without the BSP loss. Therefore, we use the square root of the Fourier amplitudes in the energy spectrum to highlight the difference following \citet{oommen2024integrating}. Although it is difficult to compare the results visually from \linkref[\del{Figure}\ins{Fig.}]{\ref{airfoil}}, we observe that the BSP loss enhances the model's ability to capture smaller scale structures given by the higher wavenumbers in the energy spectrum ($\sqrt{E(k)}$ in this case) in \linkref[\del{Figure}\ins{Fig.}]{\ref{fig:airfoil-spec}}(left). The improvement here is marginal as the model without BSP loss itself does a good job in preserving the energy spectrum of the flow field.
	\par
	To determine the performance of the BSP loss further, we compare it with a larger(as per number of parameters) state-of-the-art, Continuous Vision Transformer(CVIT) \citep{wang2024bridging} model. Due to the stochastic nature of the flow field, we compare the probability density function for the velocity values at a probe in the flow mentioned by the red dot in \linkref[\del{Figure}\ins{Fig.}]{\ref{fig:airfoil}}. In \linkref[\del{Figure}\ins{Fig.}]{\ref{fig:airfoil-spec}}(right), we observe that the UNet (trained with MSE loss) model does not preserve the probability distribution of the velocity field at the probe. However, the BSP loss improves its performance which is comparable to the approximately 60 times larger CVIT model. The UNet has a narrower distribution due to the spectral bias shifting the flow towards its mean after several rollouts. However, UNet with BSP loss has a wider distribution encompassing a  wide range of values. The BSP loss can also be implemented with the CVIT model for further comparison. Since CVIT is operated point-wise, defining the BSP loss can be challenging. The \textit{vmap} function can be used to overcome this and reshape the output to a 2D grid. Moreover, models like geo-FNO \citep{li2023fourier} can be used to extend the predictive model to non-uniform grids and BSP loss can be applied in the uniform latent dimension. We leave these paradigms for future research.
	
	\textbf{Limitations :} We remark that it is non-trivial to define the BSP loss on an unstructured grid. As demonstrated by this experiment, when applied to a problem with a non-uniform grid using interpolation, the resulting improvement is minimal. While structured grid interpolation is a somewhat ad-hoc solution to this limitation, we expect that connections to spectral techniques in numerical methods for solving partial differential equations can be leveraged for extending our proposed approach to unstructured, anisotropic, and time-varying meshes. We leave this as a task for future research.
	
	\section{Discussion}\label{Xsec14-5}
	Capturing features across a wide range of spatial and temporal scales in complex, real-world dynamical systems is a significant challenge for data-driven forecasting techniques. While recent studies have started to address the issue, they often require specialized neural architectures or end up adding substantial computational costs both during training and forecasting. To address this, we introduce a novel Binned Spectral Power (BSP) loss function that steps away from point-wise comparisons in the physical domain and instead measures differences in terms of spatial energy distributions. By applying a Fourier transform to the input fields and binning the magnitude of the Fourier coefficients by wavenumbers, we minimize discrepancies between the predicted fields and target data across multiple scales. The BSP loss offers a more balanced and efficient way to capture both large and small features without heavily modifying the model or incurring significant extra costs. Our experiments demonstrate that we can effectively reduce the spectral bias of neural networks in function approximation. We also showcase the advantages of BSP loss using challenging test cases such as turbulent flow forecasting. These results empirically show that the BSP loss function improves the ability of a neural network model to mitigate spectral bias and capture information at different scales in the data.
	
	\PrintCredit
	
	\section*{Data availability}
	Data will be made available on request. An exemplar codebase is available in \href{https://github.com/ISCLPurdue/bsp_chaos}{https://github.com/ISCLPurdue/bsp\_chaos} .
	
	\begin{coi}
		\section*{Declaration of competing interest}
		The authors declare that they have no known competing financial
		interests or personal relationships that could have appeared to
		influence the work reported in this paper.
	\end{coi}
	
	\begin{ack}
		\section*{Acknowledgements }
		This research used resources of the Argonne Leadership Computing
		Facility (ALCF), a U.S. Department of Energy (DOE) Office of Science
		user facility at Argonne National Laboratory and is based on research
		supported by the U.S. DOE Office of Science-Advanced Scientific Computing Research (ASCR) Program, under Contract No. DE-AC02-06CH11357.  RM/DC acknowledge funding support from the DOE Office of Science ASCR program (DOE-FOA-2493, PM-Dr. Steve Lee), Fusion Energy Sciences program (DOE-FOA-2905, PM-Dr. Michael Halfmoon), and computational resources from the Penn State Institute for Computational and Data Sciences. ATM acknowledges support from Artimis LDRD project at Los Alamos National Lab.
	\end{ack}
	
	\begin{appgroup}\appsection{Baseline Models and Loss Functions}\label{baselines}
		\subsection{Dilated \del{C}\ins{c}onvolutional \del{N}\ins{n}eural \del{N}\ins{n}etworks}\label{Xsec15-A.1}
		Dilated Convolutional Neural Networks (DCNNs) enhance traditional convolutional layers by introducing a dilation rate $d$ into the convolution operation. This allows the receptive field to expand exponentially without increasing the number of parameters. This architecture is used in several dynamical systems forecasting models\citep{schiff2024dyslim,chai2024overcoming,stachenfeld2021learned}.
		
		In our work we use the architecture similar to \citep{schiff2024dyslim}. It has an encoder, CNN blocks, and a decoder. The Encoder first transforms the input through two Convolutional layers with circular padding and GELU activation, ensuring smooth feature extraction. The CNN block then applies a sequence of dilated convolutions with varying dilation rates [1,2,4,8,4,2,1], allowing the network to efficiently capture both local and long-range dependencies while preserving resolution. A residual connection is added to stabilize learning and maintain input information. We employ 4 such CNN blocks. The Decoder then reconstructs the output using a couple of Convolutional layers with circular padding. The model operates recursively over multiple rollout steps, where each prediction is fed back into the network, making it particularly effective for sequence forecasting tasks.
		
		\subsection{Maximum \del{M}\ins{m}ean \del{D}\ins{d}iscrepancy (MMD) \del{L}\ins{l}oss}\label{Xsec16-A.2}
		Maximum Mean Discrepancy (MMD) used in \citet{schiff2024dyslim} is a statistical measure that quantifies the difference between two probability distributions in a reproducing kernel Hilbert space (RKHS). Given two distributions $P$ and $Q$ over a space $\mathcal{X}$, the squared MMD is defined as:
		
		\begin{equation}
			\mathrm{MMD}^2(P, Q) = \mathbb{E}_{x, x' \sim P}[k(x, x')] + \mathbb{E}_{y, y' \sim Q}[k(y, y')] - 2\mathbb{E}_{x \sim P, y \sim Q}[k(x, y)],
			\label{Xeqn21-A.1}
		\end{equation}
		where $k: \mathcal{X} \times \mathcal{X} \to \mathbb{R}$ is a positive-definite kernel. In the context of chaotic systems, MMD loss is used to match the empirical invariant measure $\mu$ with the learned distribution $\hat{\mu}$. Given observed samples $\{x_i\}_{i=1}^{N}$ and generated samples $\{\hat{x}_j\}_{j=1}^{M}$, the empirical MMD estimate is:
		\begin{equation}
			\hat{\mathrm{MMD}}^2 = \frac{1}{N^2} \sum_{i,j} k(x_i, x_j) + \frac{1}{M^2} \sum_{i,j} k(\hat{x}_i, \hat{x}_j) - \frac{2}{NM} \sum_{i,j} k(x_i, \hat{x}_j).
			\label{Xeqn22-A.2}
		\end{equation}
		Minimizing this loss ensures that the learned model captures the long-term statistical properties of the chaotic system.
		
		\subsection{Neural \del{O}\ins{o}rdinary \del{D}\ins{d}ifferential \del{E}\ins{e}quations}\label{Xsec17-A.3}
		Neural Ordinary Differential Equations (NODEs) provide a continuous-time approach to modeling dynamic systems by parameterizing the derivative of the state variable using a neural network \citep{chen2018neural}. It is described as follows:
		\begin{equation}
			\begin{aligned}
				&\frac{d\textbf{u}(t)}{dt} = \mathcal{R}(\textbf{u}(t),t,\boldsymbol{\Theta}), \quad \text{for} \quad t \in [t_0, T],
			\end{aligned}
			\label{Xeqn23-A.3}
		\end{equation}
		where \( \mathcal{R}(\textbf{u}(t),t,\boldsymbol{\Theta}) \) is a neural network parameterized by \( \boldsymbol{\Theta} \). The initial condition is given as:
		\begin{equation}
			\textbf{u}(t_0) = \textbf{u}_0.
			\label{Xeqn24-A.4}
		\end{equation}
		The solution \( \textbf{u}(t) \) is obtained by integrating the system over time using numerical solvers such as Euler's method or higher-order solvers like Runge-Kutta. In our case it can be the state of the dynamical system. The parameters \( \boldsymbol{\Theta} \) are learned by minimizing a loss function (typically MSE from ground truth) using backpropagation through the solver or with the adjoint method. Neural ODEs are particularly useful for modeling time-series data, continuous normalizing flows, and various physical systems where the dynamics are governed by differential equations \citep{chen2018neural}. Their continuous nature provides a flexible alternative to traditional discrete-layer neural networks.
		
		\subsection{Multi-step \del{P}\ins{p}enalty \del{N}\ins{n}eural ODE}\label{Xsec18-A.4}
		The Multi-step Penalty Neural ODE (MP-NODE) is formulated by \citep{chakraborty2024divide} as:
		\begin{equation}
			\begin{aligned}
				&\frac{d\textbf{u}(t)}{dt} - \mathcal{R}(\textbf{u}(t),t,\boldsymbol{\Theta}) = 0, \quad \text{for} \quad t \in [t_k, t_{k+1}) \\
				&\textbf{u}(t_k) = \textbf{u}_k^+, \quad \text{for} \quad k = 0, \ldots, n-1.
			\end{aligned}
			\label{Xeqn25-A.5}
		\end{equation}
		The corresponding loss function incorporates a penalty term and is expressed as:
		\begin{equation}\label{MP_loss_equation}
			\mathcal{L} = \mathcal{L}_{GT} + \frac{\mu}{2} \mathcal{L}_{P},
		\end{equation}
		where:
		\begin{equation}
			\mathcal{L}_{GT} = \frac{\sum_{i=1}^{N} |\textbf{u}_i - \textbf{u}_i^{true}|^2 }{2N}, \quad \mathcal{L}_{P} = \frac{\sum_{k=1}^{n-1} |\textbf{u}_k^+ - \textbf{u}_k^-|^2 }{n-1},
			\label{Xeqn27-A.7}
		\end{equation}
		represent the loss with respect to ground truth and the penalty loss enforcing continuity, respectively. For \( k = 1,2,\dots,n \), the term \( \textbf{u}_k^- \) is computed as:
		\begin{equation}
			\textbf{u}_k^- = \textbf{u}_{k-1} + \int_{t_{k-1}^+}^{t_k^-} \mathcal{R}(\textbf{u}(t),t,\boldsymbol{\Theta})\,dt.
			\label{Xeqn28-A.8}
		\end{equation}
		The penalty strength \( \mu \)(here) plays a critical role in handling local discontinuities (quantified by \( |\textbf{u}_k^+ - \textbf{u}_k^-| \)). The update strategy for \( \mu \) follows a heuristic approach, where adjustments are made based on the observed loss curves \citep{chung2022optimization}. \citet{chakraborty2024divide} show that the MP-NODE performs better for forecasting of chaotic systems.
		
		\appsection{Training Dynamics via Neural Tangent Kernel Approximation}\label{BSP theory}
		
		To understand how the Binned Spectral Power (BSP) loss potentially mitigates spectral bias, we analyze the training dynamics of Fourier modes under gradient descent. Let $\Omega=\mathbb{T}^d$ denote the $d$-dimensional unit torus (equivalently, $[0,1]^d$ with periodic boundary conditions). Let $\mathbf{f}_\theta:\Omega\to\mathbb{R}^D$ be a vector-valued neural network parameterized by $\theta\in\mathbb{R}^p$, and let $\mathbf{v}:\Omega\to\mathbb{R}^D$ be a target function. We assume $\mathbf{f}_\theta(\cdot),\mathbf{v}\in L^2(\Omega;\mathbb{R}^D)$, and that $\mathbf{f}_\theta$ is differentiable with respect to $\theta$.
		This section uses simplified definitions and reasonable assumptions following prior works on training dynamics using Neural Tangent Kernel(NTK) approximation \citep{jacot2018neural,canatar2021spectral,rahaman2019spectral}.
		For any wavevector $k \in \mathbb{Z}^d$, the Fourier coefficients of $\mathbf{f}_\theta(x)$ and $\mathbf{v}(x)$ are vectors in $\mathbb{C}^D$:
		\begin{equation}
			\hat{\mathbf{f}}_\theta(k) = \int_\Omega \mathbf{f}_\theta(x)\, e^{-2\pi i k \cdot x} \, dx, \qquad
			\hat{\mathbf{v}}(k) = \int_\Omega \mathbf{v}(x)\, e^{-2\pi i k \cdot x} \, dx.
			\label{Xeqn29-B.1}
		\end{equation}
		Each component $j \in \{1, \dots, D\}$ of these vector coefficients, $\hat{f}_{\theta,j}(k)$ and $\hat{v}_{j}(k)$, is a complex number. Since $\mathbf{f}_\theta(x)$ and $\mathbf{v}(x)$ are real-valued, their Fourier coefficients satisfy $\hat{\mathbf{f}}_\theta(-k) = \hat{\mathbf{f}}_\theta(k)^*$ and $\hat{\mathbf{v}}(-k) = \hat{\mathbf{v}}(k)^*$, where $\mathbf{z}^*$ denotes the component-wise complex conjugate of vector $\mathbf{z}$.
		
		We consider the continuous-time analogue of gradient descent:
		\begin{equation}
			\frac{d\theta}{dn} = -\nabla_\theta L(\theta),
			\label{Xeqn30-B.2}
		\end{equation}
		where $L(\theta)$ is the training loss. The evolution of the $k$-th Fourier coefficient vector $\hat{\mathbf{f}}_\theta(k)$ is then given by the chain rule, applied component-wise or using Jacobians:
		\begin{equation}
			\frac{d\hat{\mathbf{f}}_\theta(k)}{dn}
			= (\nabla_\theta \hat{\mathbf{f}}_\theta(k)) \frac{d\theta}{dn}
			= -(\nabla_\theta \hat{\mathbf{f}}_\theta(k)) \nabla_\theta L(\theta),
			\label{eq:fourier_chain_main_vector}
		\end{equation}
		where $\nabla_\theta \hat{\mathbf{f}}_\theta(k)$ is the $D \times p$ Jacobian matrix whose $(j,l)$-th entry is $\frac{\partial \hat{f}_{\theta,j}(k)}{\partial \theta_l}$.
		
		The Neural Tangent Kernel (NTK) for vector-valued outputs is a matrix-valued kernel. The $(j,m)$-th component of the NTK matrix $\hat{\mathbf{\Theta}}(k, k')$ (of size $D \times D$) is defined as \citep{canatar2021spectral}:
		\begin{equation}
			\hat{\Theta}_{jm}(k, k') := \left\langle \nabla_\theta \hat{f}_{\theta,j}(k), \nabla_\theta \hat{f}_{\theta,m}(k')^* \right\rangle = \sum_{l=1}^p \frac{\partial \hat{f}_{\theta,j}(k)}{\partial \theta_l} \frac{\partial \hat{f}_{\theta,m}(k')^*}{\partial \theta_l}.
			\label{eq:ntk_definition_fourier_matrix}
		\end{equation}
		
		In the infinite-width limit, $\hat{\mathbf{\Theta}}(k,k')$ is assumed constant during training \citep{jacot2018neural}. We further assume that, in the Fourier basis, it is approximately diagonal \citep{canatar2021spectral,rahaman2019spectral}:
		\begin{equation}
			\hat{\mathbf{\Theta}}(k,k') \approx \delta_{k,k'}\,\mathbf{\Theta}(k),
			\label{Xeqn33-B.5}
		\end{equation}
		where $\mathbf{\Theta}(k)$ is a $D\times D$ positive semidefinite matrix for each frequency $k$. A common further simplification is that $\mathbf{\Theta}(k)=\Theta(k)\mathbf{I}_D$, where $\Theta(k)\ge 0$.
		
		For real parameters $\theta$ and complex Fourier coefficients, the exact gradient-flow dynamics include both the standard NTK term and an anomalous kernel term:
		\[
		\frac{d\hat{\mathbf f}_\theta(k)}{dn}
		=
		-\sum_{k'} \hat{\mathbf\Theta}(k,k')\,\frac{\partial L}{\partial \hat{\mathbf f}_\theta(k')^*}
		-\sum_{k'} \hat{\mathbf A}(k,k')\,\overline{\frac{\partial L}{\partial \hat{\mathbf f}_\theta(k')^*}},
		\]
		where
		\[
		\hat A_{jm}(k,k')
		:=
		\sum_{l=1}^p
		\frac{\partial \hat f_{\theta,j}(k)}{\partial \theta_l}
		\frac{\partial \hat f_{\theta,m}(k')}{\partial \theta_l}.
		\]
		Under the simplifying assumption that anomalous terms are absorbed, together with the Fourier-diagonal approximation above, the training dynamics reduce to
		\begin{equation}
			\frac{d\hat{\mathbf f}_\theta(k)}{dn}
			\approx
			-\mathbf{\Theta}(k)\frac{\partial L}{\partial \hat{\mathbf f}_\theta(k)^*}.
			\label{eq:general_fourier_dynamics_vector}
		\end{equation}
		
		\subsection{Training dynamics under MSE \del{L}\ins{l}oss}
		
		The Mean Squared Error (MSE) loss for vector-valued functions is:
		\begin{equation}
			L_{\mathrm{MSE}}(\theta) = \frac{1}{2} \int_\Omega \| \mathbf{f}_\theta(x) - \mathbf{v}(x) \|_{\mathbb{R}^D}^2 dx = \frac{1}{2} \int_\Omega \sum_{j=1}^D ( f_{\theta,j}(x) - v_j(x) )^2 dx.
			\label{Xeqn35-B.7}
		\end{equation}
		Using Parseval's theorem (component-wise, assuming $|\Omega|=1$):
		\begin{equation}
			{L_{\mathrm{MSE}}(\theta) = \frac{1}{2} \sum_{k'} \sum_{j=1}^D \left| \hat{f}_{\theta,j}(k') - \hat{v}_j(k') \right|^2 = \frac{1}{2} \sum_{k'} \| \hat{\mathbf{f}}_\theta(k') - \hat{\mathbf{v}}(k') \|_{\mathbb{C}^D}^2.}
			\label{Xeqn36-B.8}
		\end{equation}
		The derivative vector $\frac{\partial L_{\mathrm{MSE}}}{\partial \hat{\mathbf{f}}_\theta(k)^*}$ has components $\frac{\partial L_{\mathrm{MSE}}}{\partial \hat{f}_{\theta,j}(k)^*} = \frac{1}{2} ( \hat{f}_{\theta,j}(k) - \hat{v}_j(k) )$. Thus:
		\begin{equation}
			\frac{\partial L_{\mathrm{MSE}}}{\partial \hat{\mathbf{f}}_\theta(k)^*} = \frac{1}{2} \left( \hat{\mathbf{f}}_\theta(k) - \hat{\mathbf{v}}(k) \right).
			\label{Xeqn37-B.9}
		\end{equation}
		Numerical prefactors arising from conjugate symmetry are absorbed into the effective kernel $\mathbf{\Theta}(k)$. Substituting into \linkref[Eq.~]{\eqref{eq:general_fourier_dynamics_vector}}:
		\begin{equation}
			\frac{d\hat{\mathbf{f}}_\theta(k)}{dn} \approx - \mathbf{\Theta}(k) \left( \hat{\mathbf{f}}_\theta(k) - \hat{\mathbf{v}}(k) \right).
			\label{Xeqn38-B.10}
		\end{equation}
		If $\mathbf{\Theta}(k) = \Theta(k)\mathbf{I}_D$:
		\begin{equation}
			\frac{d\hat{\mathbf{f}}_\theta(k)}{dn} \approx -\Theta(k) \left( \hat{\mathbf{f}}_\theta(k) - \hat{\mathbf{v}}(k) \right).
			\label{eq:mse_dynamics_final_vector}
		\end{equation}
		Each component of $\hat{\mathbf{f}}_\theta(k)$ evolves towards the corresponding component of $\hat{\mathbf{v}}(k)$, governed by the scalar rate $\Theta(k)$. Here $\Theta(k)$ is larger for lower modes which causes the spectral bias \citep{rahaman2019spectral}. However, we note that the term $\left( \hat{\mathbf{f}}_\theta(k) - \hat{\mathbf{v}}(k) \right)$ is also larger intuitively for lower modes.
		
		\subsection{Training dynamics under BSP \del{L}\ins{l}oss}
		
		For vector-valued functions, the spectral energy $E_\theta(k)$ at mode $k$ is typically defined as the sum of energies over all $D$ output dimensions:
		\begin{equation}
			E_\theta(k) := \tfrac{1}{2} \|\hat{\mathbf{f}}_\theta(k)\|_{\mathbb{C}^D}^2 = \tfrac{1}{2} \sum_{j=1}^D |\hat{f}_{\theta,j}(k)|^2 = \tfrac{1}{2} \hat{\mathbf{f}}_\theta(k)^\dagger \hat{\mathbf{f}}_\theta(k).
			\label{Xeqn40-B.12}
		\end{equation}
		Similarly for $E_v(k) := \tfrac{1}{2} \|\hat{\mathbf{v}}(k)\|_{\mathbb{C}^D}^2$. With this scalar definition of energy per mode $k$, {we define the BSP loss (without the binning for simplicity):
			\begin{equation}
				L_{\mathrm{BSP}}(\theta) = \sum_{k'} \left(1 - \frac{E_\theta(k') + \varepsilon}{E_v(k') + \varepsilon} \right)^2.
				\label{eq:bsp_loss_continuous_vector}
		\end{equation}}
		The derivative $\frac{\partial L_{\mathrm{BSP}}}{\partial E_\theta(k)}$ is as before: $\frac{\partial L_{\mathrm{BSP}}}{\partial E_\theta(k)} = -2 \frac{E_v(k) - E_\theta(k)}{(E_v(k) + \varepsilon)^2}$.
		The derivative of $E_\theta(k)$ with respect to a component $\hat{f}_{\theta,j}(k)^*$ is $\frac{\partial E_\theta(k)}{\partial \hat{f}_{\theta,j}(k)^*} = \tfrac{1}{2}\hat{f}_{\theta,j}(k)$.
		Thus, the $j$-th component of $\frac{\partial L_{\mathrm{BSP}}}{\partial \hat{\mathbf{f}}_\theta(k)^*}$ is:
		$ \frac{\partial L_{\mathrm{BSP}}}{\partial \hat{f}_{\theta,j}(k)^*} = \frac{\partial L_{\mathrm{BSP}}}{\partial E_\theta(k)} \frac{\partial E_\theta(k)}{\partial \hat{f}_{\theta,j}(k)^*} = \left( -2 \frac{E_v(k) - E_\theta(k)}{(E_v(k) + \varepsilon)^2} \right) \left( \tfrac{1}{2} \hat{f}_{\theta,j}(k) \right). $
		So, the derivative vector is:
		\begin{equation}
			\frac{\partial L_{\mathrm{BSP}}}{\partial \hat{\mathbf{f}}_\theta(k)^*} = - \frac{E_v(k) - E_\theta(k)}{(E_v(k) + \varepsilon)^2} \cdot \hat{\mathbf{f}}_\theta(k).
			\label{Xeqn42-B.14}
		\end{equation}
		Substituting into \linkref[Eq.~]{\eqref{eq:general_fourier_dynamics_vector}}:
		\begin{align}
			\frac{d\hat{\mathbf{f}}_\theta(k)}{dn}
			&\approx -\mathbf{\Theta}(k) \left( - \frac{E_v(k) - E_\theta(k)}{(E_v(k) + \varepsilon)^2} \cdot \hat{\mathbf{f}}_\theta(k) \right) \nonumber \\
			&\approx \mathbf{\Theta}(k) \frac{E_v(k) - E_\theta(k)}{(E_v(k) + \varepsilon)^2} \hat{\mathbf{f}}_\theta(k).
			\label{eq:bsp_dynamics_final_vector}
		\end{align}
		If $\mathbf{\Theta}(k) = \Theta(k)\mathbf{I}_D$, the dynamics become:
		\begin{equation}
			\frac{d\hat{\mathbf{f}}_\theta(k)}{dn} \approx \Theta(k) \frac{E_v(k) - E_\theta(k)}{(E_v(k) + \varepsilon)^2} \hat{\mathbf{f}}_\theta(k).
			\label{Xeqn43-B.16}
		\end{equation}
		In this case, all components of $\hat{\mathbf{f}}_\theta(k)$ are scaled by the factor, which depends on the square of the total energy $E_v(k)$ in mode $k$. This adaptive reweighting term, {$\frac{1}{(E_v(k) + \varepsilon)^2}$ (which is} based on the training data) helps mitigate spectral bias. {Additionally, the choice of $\lambda_i$ and $\epsilon$ terms in \linkref[Algorithm]{\ref{alg:spectral_loss}} are critical to balance the contribution of different frequency bands while maintaining stability during training.} In practice, while the BSP loss effectively enhances the learning of spectral magnitudes, the MSE loss is included to provide the essential phase-alignment signal, ensuring the model's output matches the target function and not just its power spectrum.
		
		\appsection{Hyperparameters} \label{hyperparams}
		
		\begin{table}[b]
			\tbl{Hyperparameters used for different models and datasets.}{%
				\begin{tabular}{lllll}
					\toprule
					{Setting} & {2D Turbulence} & {Airfoil} & {3D Turbulence} & {Airfoil Large} \\
					\colrule
					Model Name        & DCNN      & UNet      & UNet       & CVIT \\
					Parameters        & 1.1M      & 0.6M      & 90M        & 37M \\
					Learning Rate     & $10^{-3}$ to $10^{-5}$ & $5\times10^{-4}$ to $10^{-6}$ & $5\times10^{-4}$ to $10^{-6}$ & $10^{-3}$ to $10^{-6}$ \\
					Max Timesteps ($t$) & 4        & 5         & 3          & 1 \\
					$\gamma(t)$       & $0.9^{t-1}$ & $0.9^{t-1}$ & $0.9^{t-1}$ & NA \\
					$\mu$             & 1         & 0.1       & 1          & NA \\
					$\lambda_k$       & $k^2$     & 1         & $k^2$      & NA \\
					Optimizer         & Adam      & Adam      & Adam       & Adam \\
					Scheduler         & Cosine    & ReduceLROnPlateau & Cosine & NA \\
					Batch Size        & 32        & 32        & 8          & 32 \\
					\botrule
			\end{tabular}}
			\label{tab:hyperparams}
		\end{table}
	\end{appgroup}
	
	In this section, we declare the model hyperparameters in \linkref[Table]{\ref{tab:hyperparams}}. All model hyperparameters are kept same for both baselines and the model trained with BSP loss. The NODE and MPNODE models are used directly from \cite{chakraborty2024divide}. The hyperparameters of CVIT model is taken form \citep{wang2024bridging}. The length of trajectory used in training is started from 1 and gradually increased to Max Timesteps(t).
	
	\bibliographystyle{plainnat}

\end{document}